\documentclass[conference]{IEEEtran}
\IEEEoverridecommandlockouts
\usepackage{cite}
\usepackage{amsmath,amssymb,amsfonts}

\usepackage{graphicx}
\usepackage{subcaption}
\usepackage{textcomp}
\usepackage{xcolor}
\usepackage{hyperref} % for \url{...}
\usepackage[acronym]{glossaries}
\usepackage{algorithm}
\usepackage{algpseudocode}
\def\BibTeX{{\rm B\kern-.05em{\sc i\kern-.025em b}\kern-.08em
    T\kern-.1667em\lower.7ex\hbox{E}\kern-.125emX}}

\newacronym{ad}{AD}{Autonomous Driving}
\newacronym{gps}{GPS}{Global Positioning System}
\newacronym{imu}{IMU}{Inertial Measurement Unit}
\newacronym{coco}{COCO}{Common Objects in Context}
\newacronym{lss}{LSS}{Lift-Splat-Shoot}
\newacronym{bev}{BEV}{Bird's-Eye View}
\newacronym{av}{AV}{Autonomous Vehicle}
\newacronym{ros}{ROS}{Robot Operating System}
\newacronym{poe}{PoE}{Power over Ethernet}
\newacronym{ntp}{NTP}{Network Time Protocol}
\newacronym{nuc}{NUC}{Next Unit of Computing}
\newacronym{ap}{AP}{Average Precision}
\newacronym{ate}{ATE}{Average Translation Error}
\newacronym{ase}{ASE}{Average Scale Error}
\newacronym{aoe}{AOE}{Average Orientation Error}
\newacronym{sam}{SAM}{Segment Anything Model}
\newacronym{dds}{DDS}{Data Distribution Service}
\newacronym{ndrf}{NDRF}{Non-Driving Related Function}
\newacronym{ads}{ADS}{Automated Driving System}
\newacronym{odd}{ODD}{Operational Design Domain}
\newacronym{llm}{LLM}{Large Language Model}
\newacronym{vlm}{VLM}{Vision Language Model}
\newacronym{vla}{VLA}{Vision Language Action}
\newacronym{dl}{DL}{Deep Learning}
\newacronym{mrm}{MRM}{Minimal Risk Maneuver}
\newacronym{hmi}{HMI}{Human Machine Interface}
\newacronym{gpu}{GPU}{Graphic Processing Unit}
\newacronym{xai}{XAI}{eXplainable Artificial Intellegence}
\newacronym{sfm}{SfM}{Structure from Motion}
\newacronym{mvs}{MVS}{Multi-View Stereo}
\newacronym{yolo}{YOLO}{You Only Look Once}
\newacronym{har}{HAR}{Human Action Recognition}
\newacronym{rmse}{RMSE}{Root Mean Squared Error}
\newacronym{map}{mAP}{Mean Average Precision}
\newacronym{icp}{ICP}{Iterative Closest Point}
\newacronym{abd}{ABD}{AwareBEVDepth}

\makeatletter
\newcommand{\linebreakand}{%
  \end{@IEEEauthorhalign}
  \hfill\mbox{}\par
  \mbox{}\hfill\begin{@IEEEauthorhalign}
}
\makeatother

\title{Multi-View In-Cabin Monitoring System for Public Transport Vehicles}

\author{\IEEEauthorblockN{1\textsuperscript{st} Evgeny Gorelik}
\IEEEauthorblockA{\textit{GT-ARC, TU Berlin}\\
Berlin, Germany \\
evgeny.gorelik@campus.tu-berlin.de}
\and
\IEEEauthorblockN{2\textsuperscript{nd} Kenny Dean Karrow}
\IEEEauthorblockA{\textit{GT-ARC, TU Berlin}\\
Berlin, Germany \\
karrow@tu-berlin.de}
\and
\IEEEauthorblockN{3\textsuperscript{rd} Dr. Fikret Sivrikaya}
\IEEEauthorblockA{\textit{GT-ARC, TU Berlin}\\
Berlin, Germany \\
fikret.sivrikaya@tu-berlin.de}
\linebreakand
\IEEEauthorblockN{4\textsuperscript{th} Prof. Sahin Albayrak}
\IEEEauthorblockA{\textit{GT-ARC, TU Berlin}\\
Berlin, Germany \\
sahin.albayrak@tu-berlin.de}
\and
\IEEEauthorblockN{5\textsuperscript{th} Christian Baumann}
\IEEEauthorblockA{\textit{MAN Truck \& Bus SE}\\
Munich, Germany \\
christian.baumann@man.eu}
}

\begin{document}

\maketitle

\begin{abstract}
We introduce a multi-view in-cabin monitoring dataset for public transportation with synchronized RGB and depth images from four inward-facing cameras and a rotating LiDAR covering the vehicle interior of a digitalized and partly automated German city bus. The dataset contains 9{,}136 synchronized samples with annotations and is accompanied by a calibration and pseudo-labeling pipeline that generates 3D human pose estimates and oriented 3D bounding boxes for occupants. We further provide a nuScenes-format conversion and benchmark representative multi-view 3D detection models (e.g., Lift-Splat-Shoot and BEVFusion), supporting comparative evaluation and small-scale training of multi-view in-cabin perception models. The dataset and tools are available at \href{https://github.com/EvgenyGorelik/multiview_incabin_dataset}{https://github.com/EvgenyGorelik/multiview\_incabin\_dataset}.
\end{abstract}

% Sections (each in its own file under sections/)
\section{Introduction}\label{sec:introduction}

In recent years, autonomous driving has seen rapid progress in perception, planning, and control. In contrast, understanding what happens \emph{inside} the vehicle---who is present, where they are located, and how they move---has received comparatively less attention. In-cabin monitoring addresses this gap by enabling scene understanding within the passenger compartment.

In public transportation, in-cabin perception can support occupant counting and localization, monitoring of boarding and alighting, assistance systems for passengers with reduced mobility, and early detection of hazardous or emergency situations. These applications benefit from robust 3D understanding, but are challenging in practice due to crowding, severe occlusions, reflective surfaces, and limited viewpoints.

A key bottleneck is data. Existing in-cabin datasets are often limited in scale and typically provide only a single camera view \cite{mishra2022cabin, mishra2020intelligent, poon2022yolo, tsiktsiris2025complete, lin2024multi}. Moreover, multi-sensor calibration in buses is difficult because inward-facing cameras have only limited overlap. Covering the full cabin while reducing blind spots and handling occlusions therefore calls for a calibrated multi-view setup and synchronized multi-modal measurements.

In this work, we present a multi-view in-cabin monitoring dataset for public transportation, captured inside a digitized German city bus with synchronized RGB, stereo depth, and LiDAR. The dataset is designed as a controlled testbed for multi-view in-cabin perception under strong occlusion. Specifically, we contribute:
\begin{itemize}
    \item a multi-view, multi-modal in-cabin dataset captured in a full-scale city-bus interior, released in the nuScenes format;
    \item a practical calibration and pseudo-labeling pipeline that produces 3D pose estimates, cross-view tracks, and oriented 3D bounding boxes for occupants;
    \item baseline results for representative state-of-the-art multi-view 3D detection models on the proposed benchmark.
\end{itemize}

\section{Related Work}\label{sec:related_work}

This section reviews prior work on in-cabin datasets, multi-view and multi-modal sensing, 3D object detection, 3D human pose estimation, and pseudo-labeling for semi-supervised learning.

\subsection{In-Cabin Datasets}

Publicly available datasets are essential for training and benchmarking in-cabin monitoring systems, yet existing resources remain comparatively limited in scale and scope. Although some datasets target specific tasks such as action recognition \cite{lin2024multi}, others provide more general in-cabin annotations \cite{mishra2022cabin}.

Mishra \emph{et al.}~\cite{mishra2020intelligent} target \emph{FAV (Level~5) safety}, where no human driver is present to manage the cabin, and emphasize irregular behavior recognition. Tsiktsiris \emph{et al.}~\cite{tsiktsiris2020abnormal} focus on \emph{autonomous shuttle security} via abnormal event/action monitoring. Mishra \emph{et al.}~\cite{mishra2022cabin} highlight a \emph{privacy--forensic balance} using anonymization while maintaining evidential utility during abnormal events. Poon \emph{et al.}~\cite{poon2022yolo} target \emph{driver and occupant status} understanding using in-cabin action and object cues. Ciampi \emph{et al.}~\cite{ciampi2022bus} emphasize \emph{domain generalization} by evaluating violence detection ``in-the-wild'' across varied public-transport conditions. Lin \emph{et al.}~\cite{lin2024multi} study \emph{dangerous driver behavior} with RGB-D sensing and multi-task supervision. Lin \& Tseng~\cite{lin2024bushar} focus on an \emph{overhead (top-down) perspective} for crowded buses to reduce occlusions, which complicates pose-based reasoning. Finally, Tsiktsiris \emph{et al.}~\cite{tsiktsiris2024overhead, tsiktsiris2025complete} stress \emph{multimodal sensor fusion} (RGB, depth, audio) to capture events that may be visually occluded.

Table~\ref{tab:in_cabin_datasets_comparison} summarizes representative datasets by sensing modality and supported tasks and annotations.

\begin{table*}[ht!]
    \centering
    \small
    \setlength{\tabcolsep}{4pt}
    \renewcommand{\arraystretch}{1.15}
    \begin{tabular}{lcccccccc}
        \hline\hline
        \textbf{Dataset} & \textbf{RGB} & \textbf{Depth} & \textbf{LiDAR} & \textbf{Audio} & \textbf{Action} & \textbf{Pose} & \textbf{Obj.} & \textbf{Pub. Av.}\\
        \hline
        Mishra \emph{et al.} (2020)~\cite{mishra2020intelligent} & $\checkmark$ & -- & -- & -- & $\checkmark$ & -- & $\checkmark$ & -- \\
        Tsiktsiris \emph{et al.} (2020)~\cite{tsiktsiris2020abnormal} & $\checkmark$ & -- & -- & -- & $\checkmark$ & $\checkmark$ & $\checkmark$ & -- \\
        Mishra \emph{et al.} (2022)~\cite{mishra2022cabin} & $\checkmark$ & -- & -- & -- & $\checkmark$ & -- & $\checkmark$ & -- \\
        Poon \emph{et al.} (2022)~\cite{poon2022yolo} & $\checkmark$ & -- & -- & -- & $\checkmark$ & -- & $\checkmark$ & -- \\
        Ciampi \emph{et al.} (2022)~\cite{ciampi2022bus} & $\checkmark$ & -- & -- & -- & $\checkmark$ & -- & -- & $\checkmark$ \\
        Lin \emph{et al.} (2021/2023)~\cite{lin2024multi} & $\checkmark$ & $\checkmark$ & -- & -- & $\checkmark$ & $\checkmark$ & $\checkmark$ & -- \\
        Lin \& Tseng (2024)~\cite{lin2024bushar} & $\checkmark$ & -- & -- & -- & $\checkmark$ & -- & $\checkmark$ & -- \\
        Tsiktsiris \emph{et al.} (2024)~\cite{tsiktsiris2024overhead} & $\checkmark$ & $\checkmark$ & -- & $\checkmark$ & $\checkmark$ & -- & -- & -- \\
        Tsiktsiris \emph{et al.} (2025)~\cite{tsiktsiris2025complete} & $\checkmark$ & $\checkmark$ & -- & $\checkmark$ & $\checkmark$ & $\checkmark$ & $\checkmark$ & -- \\
        \textbf{Ours (this work)} & $\checkmark$ & $\checkmark$ & $\checkmark$ & -- & $\checkmark$ & $\checkmark$ & $\checkmark$ & $\checkmark$ \\
        \hline\hline
    \end{tabular}
    \caption{Comparison of representative in-cabin datasets by sensing modality and supported tasks/annotations. ``Action'' includes action/activity recognition; ``Pose'' includes 2D keypoints; ``Obj.'' includes object detection. The last column denotes whether the dataset is publicly available.}
    \label{tab:in_cabin_datasets_comparison}
\end{table*}

\subsection{Multi-View Datasets}

Most of the above in-cabin datasets rely on a single camera view, which restricts their applicability to larger vehicles and limits robustness under occlusion. In contrast, a multi-camera setup can increase coverage of the passenger compartment and reduce blind spots.

In autonomous driving, multi-modal datasets are a standard foundation for perception research. They typically provide synchronized measurements from multiple cameras, LiDAR, RADAR, and localization sensors such as \acrshort{gps} and \acrshort{imu}. Representative examples include KITTI \cite{geiger2013vision}, nuScenes \cite{caesar2020nuscenes}, the Waymo Open Dataset \cite{sun2020scalability}, and Argoverse/Argoverse~2 \cite{chang2019argoverse, wilson2023argoverse}. However, these benchmarks are built around outward-facing, environment-centric sensor configurations and are therefore not directly transferable to in-cabin monitoring.

Multi-sensor capture is also relatively uncommon in human pose estimation, where many benchmarks focus on monocular images, such as MPII \cite{andriluka20142d} and \acrshort{coco} \cite{lin2014microsoft}. Multi-view datasets such as CMU Panoptic Studio \cite{joo2015panoptic} and Human3.6M \cite{ionescu2013human3} provide high-quality supervision, but are collected in controlled environments and do not reflect the geometry, viewpoints, and occlusions found in vehicle cabins. To the best of our knowledge there are no publicly available multi-view, multi-modal in-cabin datasets with 3D human pose annotations.

\subsection{3D Object Detection}

3D object detection focuses on localizing and classifying objects directly in 3D space, typically by estimating an oriented 3D bounding box for each instance \cite{mao20233d}. In autonomous driving, the dominant inputs are cameras and LiDAR: camera-based methods benefit from rich appearance cues but must infer depth, while LiDAR-based methods provide more direct geometric structure at the cost of sparse point measurements and higher sensor cost. Many modern approaches are designed to bridge these trade-offs through improved depth reasoning, multi-view geometry, or multi-modal fusion \cite{ming2021deep, philion2020lift, li2024bevformer, liu2023bevfusion}.

\subsubsection{Camera-based 3D Object Detection}

Camera-only 3D detection is challenging because depth is not observed directly and must be inferred, for example via monocular depth cues or explicit depth estimation \cite{ming2021deep}. Due to this inherent ambiguity, monocular methods often lag behind LiDAR-based approaches in accuracy \cite{valverde2025survey}, but remain attractive because cameras are inexpensive and widely deployed.

\subsubsection{Multi-View 3D Object Detection}

Multi-view methods reduce depth ambiguity by aggregating information across camera views, often improving performance toward LiDAR-based baselines \cite{ming2021deep}. Approaches such as \acrfull{lss} \cite{philion2020lift} and BEVFormer \cite{li2024bevformer} lift and project image features into a common \acrshort{bev} representation using calibrated camera geometry. Working in \acrshort{bev} also enables multi-modal fusion; for instance, BEVFusion \cite{liu2023bevfusion} combines camera and LiDAR features in a shared representation.

\subsection{3D Human Pose Estimation}

While 2D pose estimation has matured substantially (e.g., \acrshort{yolo}-Pose \cite{maji2022yolo}), 3D pose estimation remains limited by depth ambiguity and by the availability of large-scale datasets with accurate 3D supervision. Methods such as SelfPose3D \cite{Srivastav_2024_CVPR} and PoseFormer \cite{zheng20213d} are often trained and evaluated on CMU Panoptic Studio \cite{joo2015panoptic}, but transferring to in-cabin setups is non-trivial due to different viewpoints and sensing configurations. Recent work such as \acrshort{sam}~3 Body 3D \cite{yang2026sam} explores monocular inference based on \acrshort{sam}~3 \cite{carion2025sam}.

\subsection{Pseudo-Labeling and Semi-Supervised Learning}

Pseudo-labeling is widely used to exploit unlabeled data in semi-supervised settings. A common strategy is to use a computationally expensive teacher model trained on labeled data to generate predictions on unlabeled data, and then train a smaller student model on these pseudo labels \cite{lee2013pseudo}. This idea is closely related to knowledge distillation \cite{hinton2015distilling} and has been applied to large-scale image classification \cite{yalniz2019billion, xie2020self} and domain adaptation \cite{french2017self}.

\section{Dataset}\label{sec:dataset}

We present a multi-view in-cabin monitoring dataset captured inside a digitized German city bus. The dataset consists of synchronized RGB and depth images from four inward-facing cameras and a rotating LiDAR, recorded in a stationary vehicle.

\paragraph{Scene content}
The dataset contains scenes with 1--2 occupants performing a range of in-cabin activities, including entering or alighting the vehicle, walking through the cabin, and sitting or standing, as well as behaviors such as drinking (alcoholic or non-alcoholic) beverages and smoking. It further covers safety- and security-critical events such as vandalism, aggression, fighting, and armed robbery, and includes accessibility scenarios such as wheelchair entry.

\paragraph{Sensors and synchronization}
Each camera is an Intel RealSense D435i connected to a Raspberry~Pi. RGB streams run at 15~Hz and the LiDAR (Ouster OS0-128) runs at 10~Hz. We form one \emph{sample} by pairing each LiDAR frame with the temporally closest frames from all cameras using timestamps.

\paragraph{Scale}
Overall, we collect 10{,}034 synchronized samples (LiDAR-timed), with corresponding multi-view images, with 9{,}136 of them containing valid annotations. We additionally store stereo depth from the RealSense devices from all cameras.

\paragraph{Calibration}
We estimate the multi-camera extrinsics by reconstructing the cabin with COLMAP with scattered ArUco markers present in the scene, scaling and orienting the reconstruction into a metric frame from the known origin-marker size, and registering each camera against its ArUco corner correspondences. We then align LiDAR to the COLMAP frame via ICP registration.

\paragraph{Annotations and tasks}
We provide pseudo-labels for people in the cabin: 2D masks and boxes per view, reconstructed 3D meshes/skeletons per view, cross-view aggregated tracks, and derived oriented 3D bounding boxes. In addition, we provide hand-labeled action annotations for the occupants, covering both state categories (\emph{sit}, \emph{stand}, \emph{hold on}) and action categories (e.g., \emph{drinking}, \emph{smoking}, \emph{punching}, \emph{vandalizing}). The dataset is intended for benchmarking multi-view, camera-based 3D detection and for small-scale training.

\paragraph{Public release}
The dataset and codebase (including the nuScenes-format conversion used in our experiments) will be released publicly upon publication.
\section{Methodology}\label{sec:methodology}

This section describes our pipeline for collecting and processing the proposed multi-view in-cabin dataset, including data acquisition, sensor calibration, pseudo-label generation, and 3D bounding-box extraction. The dataset is created using the following steps:
\begin{enumerate}
    \item \textbf{Data capture:} Record calibrated and synchronized camera and LiDAR data.
    \item \textbf{2D box extraction:} Run a 2D detector to obtain candidate person bounding boxes.
    \item \textbf{Manual filtering:} Remove obvious false positives and inconsistent detections.
    \item \textbf{3D pose estimation:} Estimate a 3D human mesh and skeleton for each detection.
    \item \textbf{Multi-view track aggregation:} Associate tracks across views and time.
    \item \textbf{3D box extraction:} Derive oriented 3D bounding boxes from the aggregated poses.
\end{enumerate}

\begin{figure}[ht!]
    \centering
    \includegraphics[width=\linewidth]{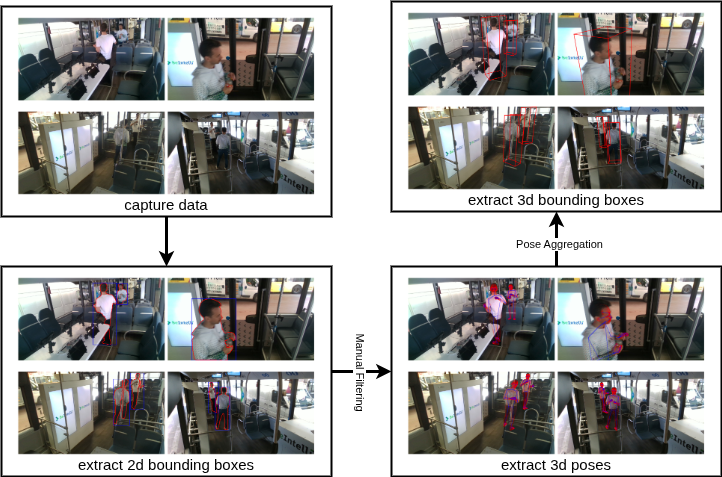}

    \caption{In-cabin monitoring dataset creation pipeline}
    \label{fig:methodology}
\end{figure}

The resulting dataset contains synchronized multi-view images, LiDAR measurements, and pseudo-labels for people in the vehicle cabin, enabling training and evaluation of multi-view 3D perception methods. Figure~\ref{fig:methodology} provides an overview of the pipeline.

\subsection{Data Collection}\label{subsec:data_collection}

\begin{figure}[!ht]
    \centering
    \includegraphics[width=0.5\linewidth]{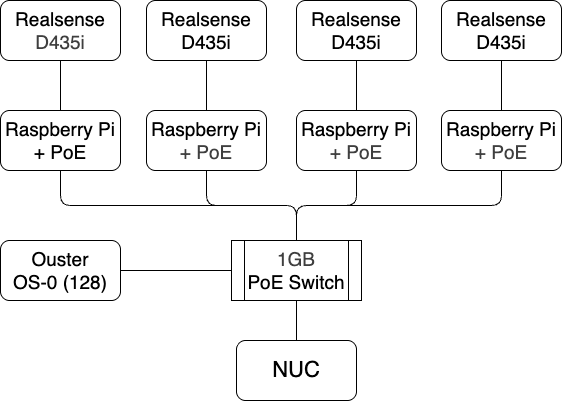}
    
    \caption{Data collection setup with Raspberry Pi devices and RealSense D435i camera.}
    \label{fig:setup}
\end{figure}

We use a distributed capture setup based on Raspberry~Pi devices, each connected to an Intel RealSense D435i camera, to record multi-view video inside the vehicle cabin. The devices are powered via a \acrfull{poe} switch and stream data to an \acrshort{nuc}~11 running \acrshort{ros}~2. Camera clocks are synchronized via \acrshort{ntp} with the \acrshort{nuc}~11 acting as the central time server. We use Cyclone \acrshort{dds} as middleware to enable reliable, low-latency data transfer. The \acrshort{poe} switch provides 1~Gbps bandwidth, which is sufficient to stream compressed high-resolution images from each Raspberry~Pi to the \acrshort{nuc}~11. Figure~\ref{fig:setup} illustrates the overall setup.

\begin{figure}[!ht]
    \centering
    \includegraphics[width=1.0\linewidth]{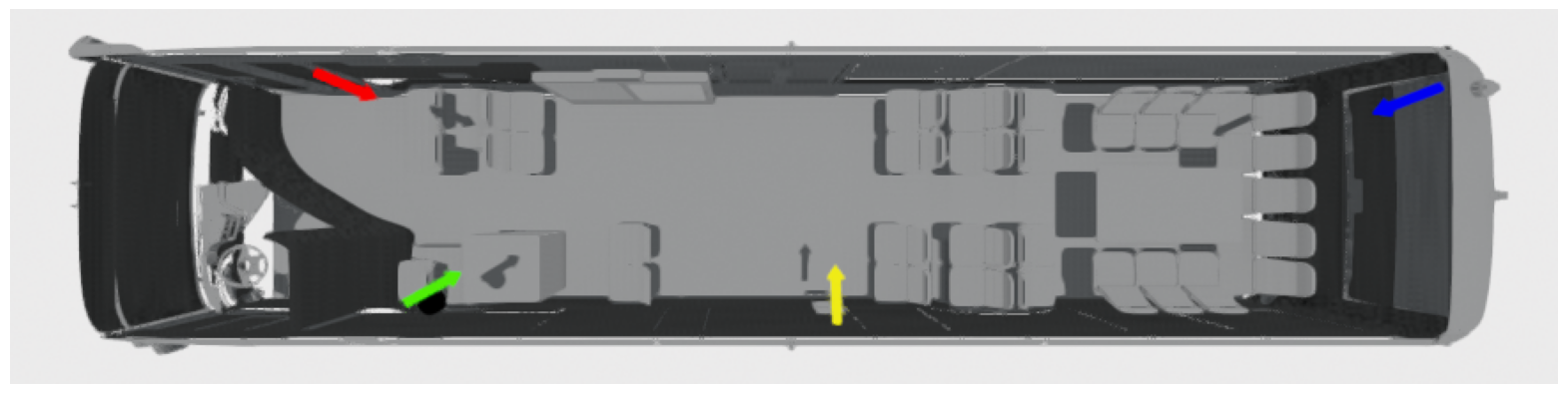}
    
    \caption{Sensor positions within the vehicle cabin.}
    \label{fig:cam_pos}
\end{figure}

\begin{figure}[ht!]
    \centering
    \includegraphics[width=1\linewidth]{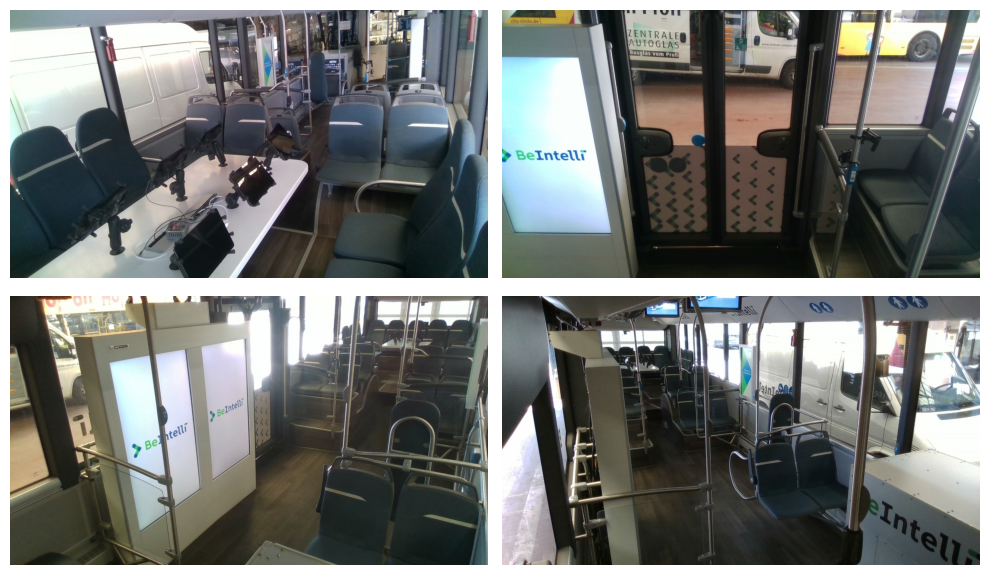}
    \caption{Sample frames from the four cameras showing different views of the vehicle interior.}
    \label{fig:cam_views}
\end{figure}

The cameras are positioned to capture complementary views of the passenger compartment, providing broad coverage for subsequent in-cabin analysis. Figure~\ref{fig:cam_pos} shows the resulting sensor layout. We intentionally do not cover the driver cabin in order to focus on passenger monitoring. Figure~\ref{fig:cam_views} shows example frames from the four cameras.

\begin{figure}[ht!]
    \centering
    \includegraphics[width=1\linewidth]{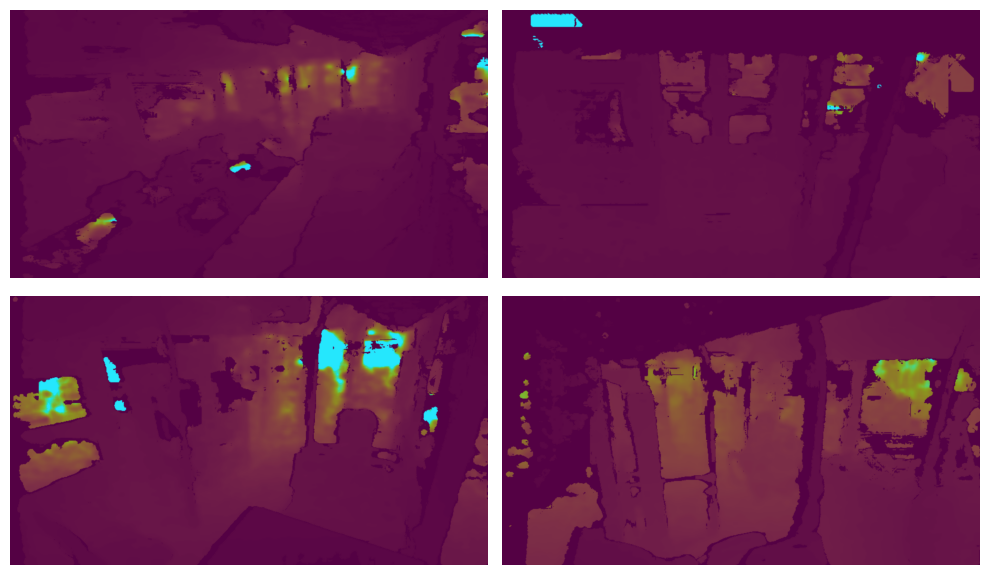}
    \caption{Samples from RealSense D435i depth estimation}
    \label{fig:cam_depth}
\end{figure}

The Intel RealSense D435i provides a stereo-based depth estimation module that outputs metric depth measurements, with best performance at distances of approximately 0.3\,m to 3.0\,m. Example depth maps produced by this module are shown in Figure~\ref{fig:cam_depth}.

\begin{figure}[ht]
    \centering
    \includegraphics[width=\linewidth]{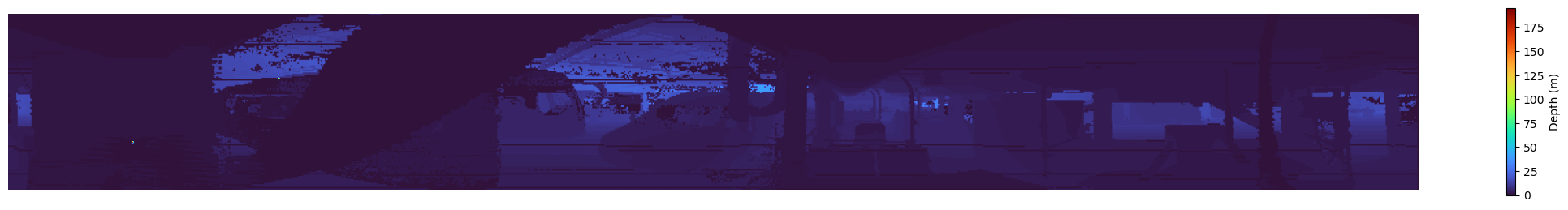}
    \caption{Sample depth image from the Ouster OS0-128 LiDAR.}
    \label{fig:lidar_depth}
\end{figure}

Additionally, we mount an Ouster OS0-128 LiDAR inside the vehicle (black cylinder in Figure~\ref{fig:cam_pos}) to provide depth measurements. Figure~\ref{fig:lidar_depth} shows an example depth visualization.

During data collection, we record synchronized RGB images from all cameras together with the LiDAR point cloud. For the cameras we also store the stereo depth produced by the RealSense D435i. Table~\ref{tab:recorded_data} summarizes the recorded modalities and the number of frames per sensor.

\begin{table}[ht]
    \centering
    \begin{tabular}{llll}
        \hline
        Device / Sensor & Location & Datatype & Num. samples \\
        \hline
        \textcolor{yellow}{Camera 1} & Center Left & RGB & 12862 \\
        \textcolor{yellow}{Camera 1} & Center Left & Depth & 13747 \\
        \textcolor{red}{Camera 2} & Front Right & RGB & 12992 \\
        \textcolor{red}{Camera 2} & Front Right & Depth & 13062 \\
        \textcolor{green}{Camera 3} & Front Left & RGB & 12890 \\
        \textcolor{green}{Camera 3} & Front Left & Depth & 12404 \\
        \textcolor{blue}{Camera 4} & Back Right & RGB & 11821 \\
        \textcolor{blue}{Camera 4} & Back Right & Depth & 12414 \\
        LiDAR & Front Center & Point cloud & 10034 \\
        \hline
    \end{tabular}
    
    \caption{Recorded data modalities by device and number of synchronized samples.}
    \label{tab:recorded_data}
\end{table}

The LiDAR operates at 10~Hz, whereas the cameras run at 15~Hz. We temporally align modalities by pairing each LiDAR frame with the closest camera frames in time using timestamps. We refer to one such aligned set (multi-view images plus the corresponding LiDAR frame) as a \textit{data sample}. This terminology matches other multi-modal datasets such as Waymo and KITTI \cite{sun2020scalability, geiger2013vision}; in contrast, nuScenes \cite{caesar2020nuscenes} uses a 2~Hz annotated sample rate and represents intermediate frames as sweeps.

Overall, we collect 10,034 synchronized samples consisting of multi-view camera images and LiDAR depth data.

\subsection{Sensor Calibration}\label{subsec:sensor_calibration}

To enable accurate sensor fusion, we require both intrinsic and extrinsic calibration for the cameras and the LiDAR. For intrinsics (focal length, principal point, and distortion), we use the manufacturer-provided parameters of the RealSense D435i, which we found sufficient in practice.

Extrinsic calibration aligns all sensors into a common coordinate system. While multi-camera and camera--LiDAR calibration are well studied in autonomous driving, many established methods assume outward-facing sensors with overlapping fields of view. In our setting, the cameras face inward and their views overlap only minimally (Figure~\ref{fig:cam_pos}), which makes standard target-based \cite{huang2020improvements} and targetless \cite{koide2023general, yan2022opencalib} approaches difficult to apply.

We therefore perform extrinsic calibration in two stages. We scatter ArUco markers of known size throughout the scene, ensuring that each camera to be calibrated observes several markers. One additional marker is deliberately placed at the vehicle’s lateral centerline, aligned with the longitudinal axis, and is used as the origin (base link) for all coordinate transformations. Because this marker is mounted level with the bus floor, it provides a ground-aligned reference frame and removes the need for explicit ground-plane estimation.

In the first stage, we reconstruct the cabin geometry. We record a handheld video of the interior, extract its frames, bin them temporally, and keep the sharpest frame per bin according to a Laplacian-variance sharpness measure. We then run COLMAP \gls{sfm} with a sequential feature matcher to obtain a sparse reconstruction of the cabin, which we densify with COLMAP \gls{mvs} \cite{schoenberger2016sfm, leibe_pixelwise_2016}.

In the second stage, we bring this reconstruction into the metric reference frame. We detect the corners of every marker in the reconstruction and, using the known side length of the origin marker, estimate a similarity transform that maps the middle of the marker to the origin, fixing both the metric scale and the orientation of the scene \cite{meyer2023cherrypicker}. Applying this transform to the full reconstruction yields a metric model. We then register each camera by minimizing a Huber loss on the distance between its detected ArUco corners and the corresponding reconstructed corners. Finally, we crop and statistically clean the dense point cloud.

% \begin{table}[ht!]
% \centering
% \begin{tabular}{r|cccc|c}
% \hline
% Marker & Edge 0--1 & Edge 1--2 & Edge 2--3 & Edge 3--0 & Avg \\
% \hline
% 0 & 0.17034 & 0.16920 & 0.17110 & 0.16937 & 0.17000 \\
% 1 & 0.17758 & 0.17470 & 0.17689 & 0.17415 & 0.17583 \\
% 2 & 0.17576 & 0.17602 & 0.17485 & 0.17614 & 0.17569 \\
% 3 & 0.17921 & 0.17699 & 0.17673 & 0.17727 & 0.17755 \\
% 4 & 0.18035 & 0.17385 & 0.18003 & 0.17352 & 0.17694 \\
% 5 & 0.17414 & 0.17513 & 0.17814 & 0.17504 & 0.17561 \\
% 6 & 0.17183 & 0.17290 & 0.17287 & 0.17289 & 0.17262 \\
% 7 & 0.17406 & 0.17221 & 0.17687 & 0.17146 & 0.17365 \\
% 8 & 0.17570 & 0.17309 & 0.17458 & 0.17455 & 0.17448 \\
% 9 & 0.16729 & 0.18469 & 0.17973 & 0.17624 & 0.17699 \\
% 10 & 0.20853 & 0.17144 & 0.19445 & 0.17488 & 0.18732 \\
% $11^{\dagger}$ & 0.22097 & 0.18130 & 0.20768 & 0.18350 & 0.19836 \\
% 12 & 0.17179 & 0.17135 & 0.17231 & 0.17225 & 0.17192 \\
% 13 & 0.17427 & 0.17838 & 0.18150 & 0.17743 & 0.17789 \\
% 14 & 0.17661 & 0.17283 & 0.17793 & 0.17367 & 0.17526 \\
% \hline
% \multicolumn{5}{r|}{\textbf{Mean edge length (excl.\ 11)}} & \textbf{0.17584} \\
% \hline
% \end{tabular}
% \caption{Edge lengths per marker. $^{\dagger}$Marker~11 is excluded from calibration due to insufficient visibility in the cameras and therefore the average is computed over the remaining markers.}
% \label{tab:edge-lengths}
% \end{table}
\begin{table}[t]
\centering
\begin{tabular}{r r @{\hskip 1.5em} r r @{\hskip 1.5em} r r}
\hline
M-ID & Avg & M-ID & Avg & M-ID & Avg \\
\hline
0 & \textbf{0.17000} & 5 & 0.17561 & 10 & 0.18732 \\
1 & 0.17583 & 6 & 0.17262 & $11^{\dagger}$ & 0.19836 \\
2 & 0.17569 & 7 & 0.17365 & 12 & 0.17192 \\
3 & 0.17755 & 8 & 0.17448 & 13 & 0.17789 \\
4 & 0.17694 & 9 & 0.17699 & 14 & 0.17526 \\
\hline
\multicolumn{6}{r}{\textbf{Mean edge length (excl.\ 11): 0.17584}} \\
\hline
\end{tabular}
\caption{Average edge length for each marker ID (in meters). $^{\dagger}$Marker~11 is excluded
from calibration due to insufficient visibility in the cameras. The mean is
computed over the remaining markers.}
\label{tab:edge-lengths}
\end{table}

We measure the edge length of the marker to be 0.1700\,m and validate the metric consistency of the reconstruction by computing the average reconstructed edge length for each marker. As shown in Table~\ref{tab:edge-lengths}, the edge lengths are recovered accurately for most markers, with an overall mean reconstructed edge length of 0.17584\,m and a fitted mean residual of 0.67\,mm at the four corners of the origin marker.

\begin{figure}[ht!]
    \includegraphics[width=\linewidth]{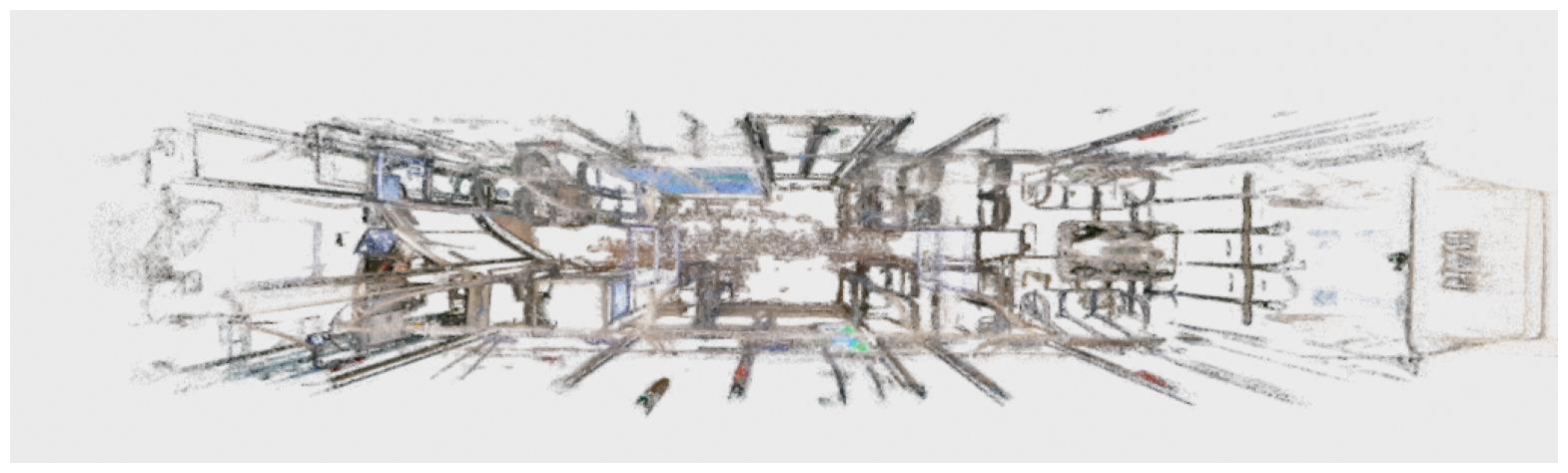}
    
    \caption{Densified and downsampled 3D point cloud reconstructed using COLMAP}
    \label{fig:colmap_reconstruction}
\end{figure}

Finally, we estimate the LiDAR extrinsics by aligning the LiDAR point cloud with the COLMAP reconstruction using an \acrshort{icp}-based registration \cite{chen1992object}, achieving a fitness score of 0.0869 and a \acrshort{rmse} of 0.0071. This provides the LiDAR pose in the same coordinate system as the cameras.

\begin{figure}[ht!]
    \centering
    \includegraphics[width=1\linewidth]{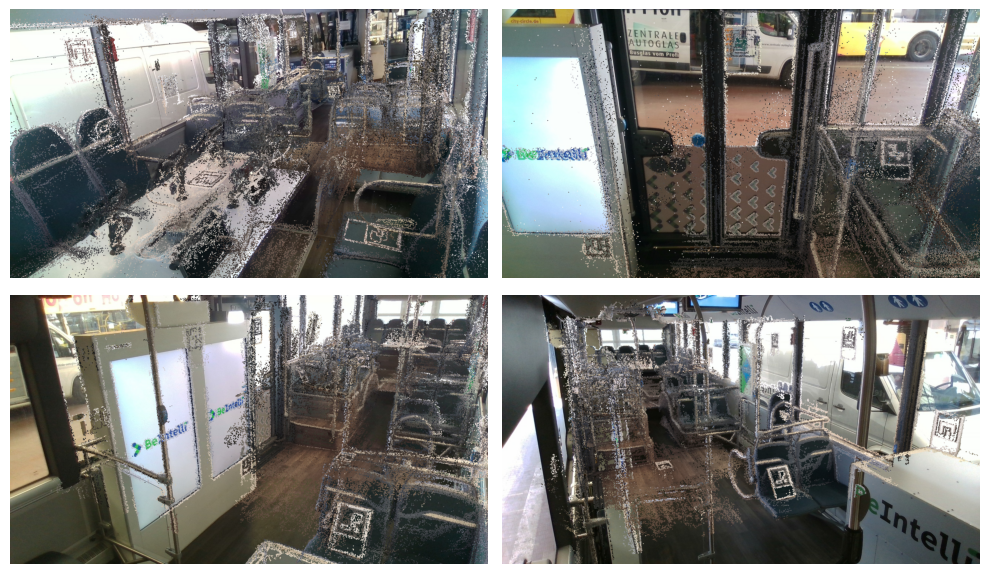}
    
    \caption{Reconstructed point cloud from COLMAP reprojected into camera views}
    \label{fig:cam_colmap}
\end{figure}

\begin{figure}[ht!]
    \centering
    \includegraphics[width=1\linewidth]{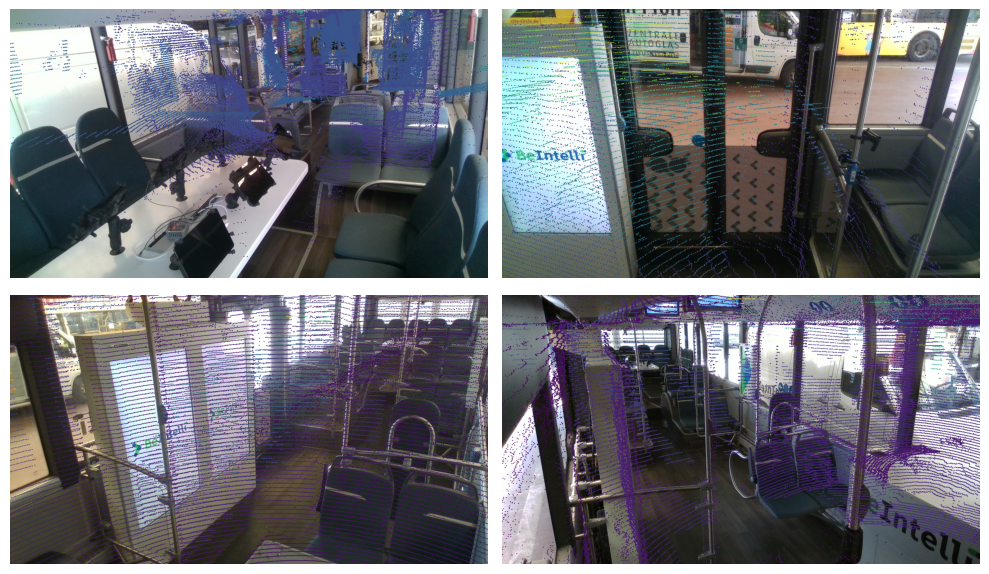}
    
    \caption{LiDAR scan after ICP reprojected into camera views}
    \label{fig:cam_lidar}
\end{figure}

Figure~\ref{fig:cam_colmap} visualizes the COLMAP reconstruction and Figure~\ref{fig:cam_lidar} visualizes the LiDAR point cloud after alignment. 

\subsection{3D Human Pose Estimation}

\begin{figure}[ht!]
    \centering
    \includegraphics[width=\linewidth]{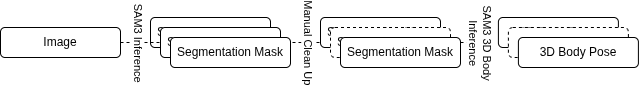}

    \caption{Overview of the pseudo-label generation process for a single camera.}
    \label{fig:autolabeling_single}
\end{figure}

We implement a pseudo-labeling pipeline that produces 3D bounding boxes for people inside vehicles. Specifically, we use \acrshort{sam}~3 \cite{carion2025sam} to obtain segmentation masks and 2D bounding boxes, and \acrshort{sam}~3 Body 3D \cite{yang2026sam} to estimate 3D human meshes and poses. Figure~\ref{fig:autolabeling_single} illustrates the per-camera pseudo-label generation process. We apply the same procedure to all four cameras, yielding a set of per-view 3D poses for each time step.

Table~\ref{tab:bb2d_stats} reports the number of manually filtered 2D bounding boxes for each camera. Most rejected detections correspond to symbols or pictures of humans that are falsely labeled as real people; since these artifacts are stationary, we automatically remove them using a fixed region-of-interest crop. As a result, only a small subset of cases---primarily reflections and duplicate detections---required manual filtering.

\begin{table}[ht!]
    \centering
    \begin{tabular}{lrr}
        \hline
        \textbf{Camera} & \textbf{Valid boxes} & \textbf{Filtered boxes} \\
        \hline
        Back right  & 14645 & 366  \\
        Front left  & 15688 & 486  \\
        Front right & 13824 & 1345 \\
        Center left & 8475  & 14265 \\
        \hline
    \end{tabular}
    \caption{Per-camera statistics of 2D bounding boxes used for 3D pose estimation.}
    \label{tab:bb2d_stats}
\end{table}

\begin{figure}[ht!]
    \centering
    \includegraphics[width=\linewidth]{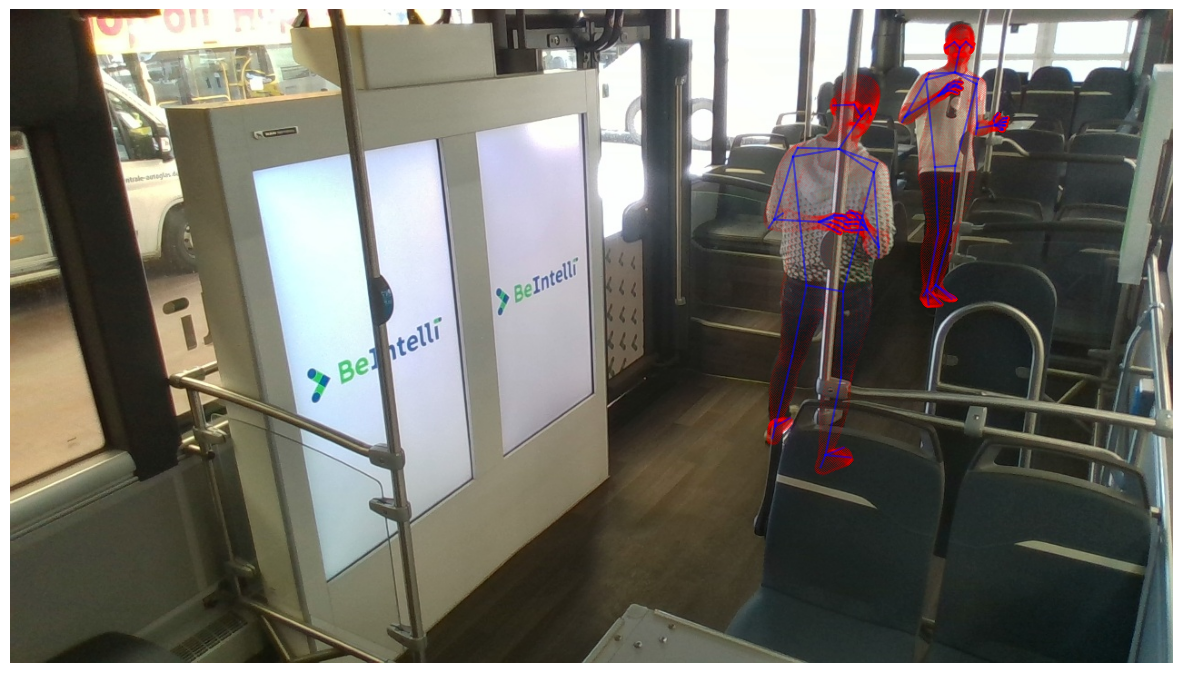}

    \caption{Joints (blue) and 3D mesh generated (red) from \acrshort{sam}~3 Body 3D inference}
    \label{fig:body3d_demo}
\end{figure}

\acrshort{sam}~3 Body 3D follows a three-stage procedure: it (i) detects 2D bounding boxes, (ii) estimates camera intrinsics, and (iii) reconstructs a 3D human mesh (and corresponding 3D skeleton) for each detection. In our pipeline, we bypass the first two stages by providing the image, the manually filtered 2D boxes, and the known camera intrinsics as input to the final stage. The output is a 3D mesh and skeleton in the camera coordinate system (Figure~\ref{fig:body3d_demo}).

\subsection{Multi-View Track Aggregation}

\begin{figure}[ht!]
    \centering
    \includegraphics[width=\linewidth]{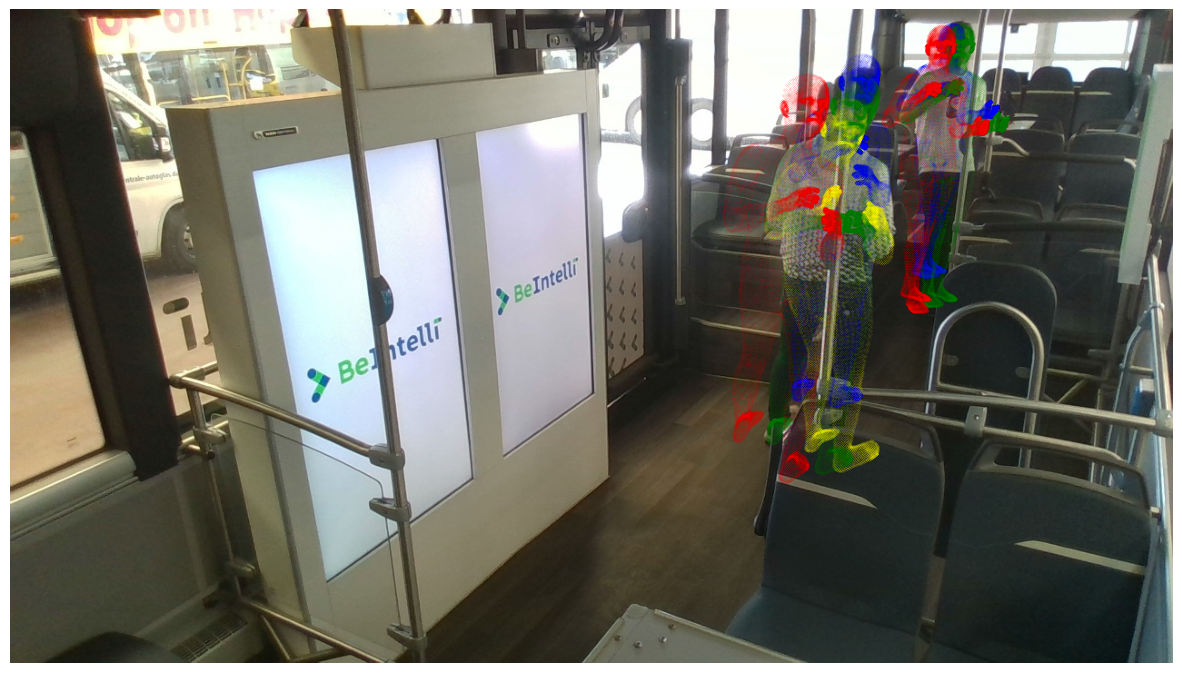}

    \caption{Example of the generated 3D meshes from \textcolor{yellow}{Camera 1}, \textcolor{red}{Camera 2}, \textcolor{green}{Camera 3} and \textcolor{blue}{Camera 4} for all people in the scene}
    \label{fig:mesh_intersection_demo}
\end{figure}

The above pipeline produces at most one 3D mesh per person and time step \,\textit{per view}. Since views overlap, the \acrshort{sam}~3 Body 3D model can generate multiple meshes for the same individual from different cameras (see Figure~\ref{fig:mesh_intersection_demo}). To enable quality control and consistent association, we first construct per-camera temporal tracks by applying ByteTrack~\cite{zhang2022bytetrack} to the \acrshort{sam}~3 2D bounding boxes. For each camera $c$, this yields a set of tracks
\begin{equation}
    \mathcal{T}^c =\left\{T_1^c, T_2^c, \ldots, T_n^c\,\middle|\,T_i^c = (\mathbf{c}, \mathbf{s}, id, \tau)\right\}.
\end{equation}
Here, $T_i^c$ is the $i$-th track in camera $c$ with bounding-box center $\mathbf{c}$ and size $\mathbf{s}$, a unique track identifier $id$, and timestamp $\tau$. We treat these per-camera track IDs as the atomic units to be associated across views. \par
We then associate each tracked detection with its \acrshort{sam}~3 Body 3D output, representing the resulting 3D pose (in the global base frame) as a matrix of $J=70$ joints:
\begin{equation}
    \mathcal{H}^c = \left\{(T^c_i, \mathbf{P}^c_i)\,\middle|\,T^c_i\in\mathcal{T}^c,\;\mathbf{P}^c_i \in \mathbb{R}^{3\times J}\right\}.
\end{equation}
These pairings provide an estimated pose for each tracked detection from each camera, which is then used for multi-view pose aggregation.

\begin{figure}[ht!]
    \centering
    \begin{subfigure}{0.48\linewidth}
        \includegraphics[width=\linewidth]{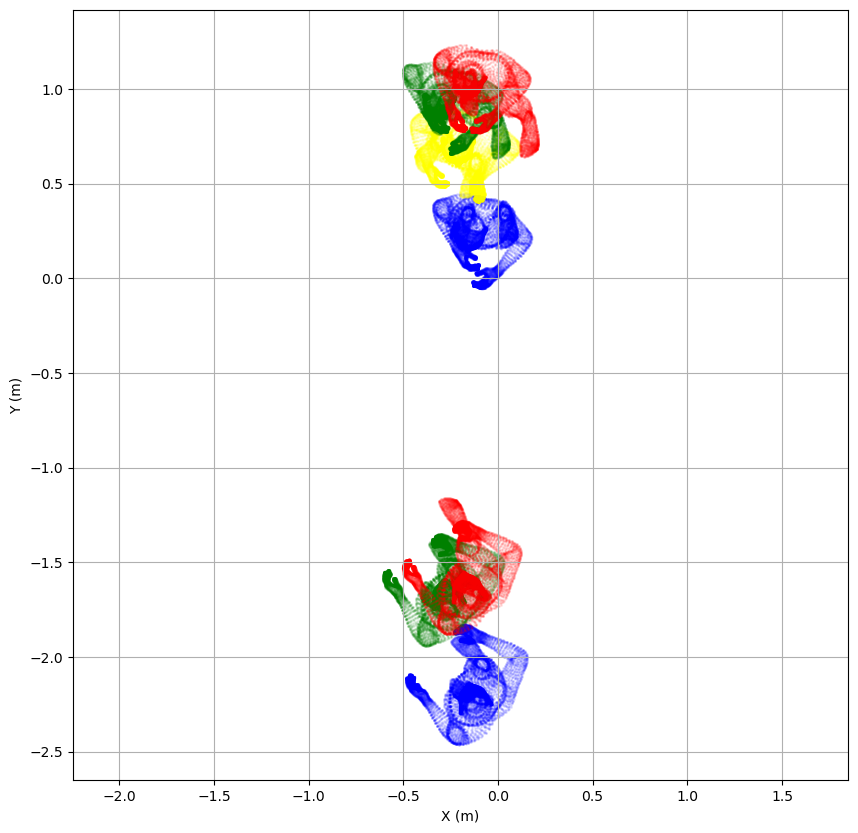}
        \label{fig:bev_multiview}
    \end{subfigure}
    \begin{subfigure}{0.48\linewidth}
        \includegraphics[width=\linewidth]{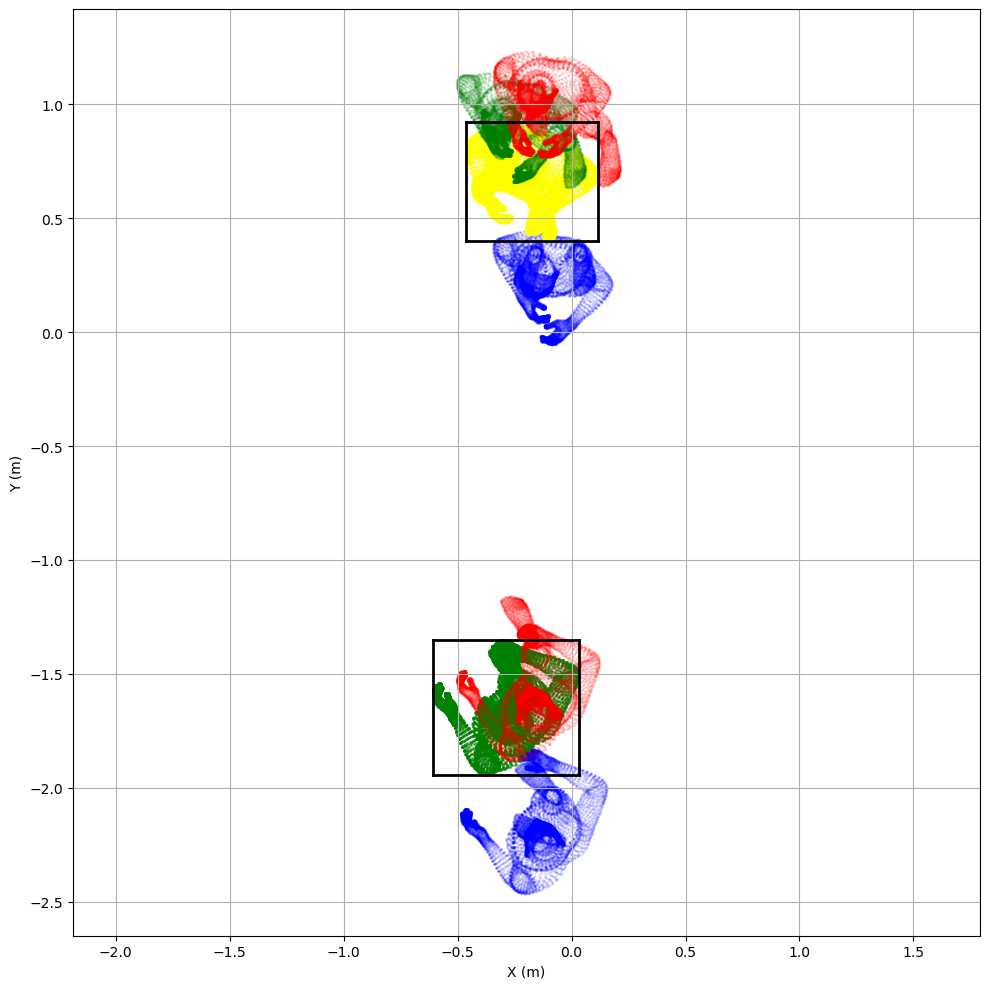}
        \label{fig:bev_multiview_select}
    \end{subfigure}

    \caption{\acrshort{bev} projection of poses (left) and median pose selection (right)}
    \label{fig:multi_view_cluster}
\end{figure}

We observe that, for a single person, the \acrshort{sam}3 Body 3D outputs from different cameras can disagree by up to $\approx 1\,\mathrm{m}$ in the \acrshort{bev} plane (Figure~\ref{fig:multi_view_cluster}). We therefore use the following three-step multi-view pose aggregation:
\begin{itemize}
    \item \textbf{Camera-constrained clustering:} cluster the per-view pose hypotheses while enforcing that a cluster contains at most one detection per camera.
    \item \textbf{Median pose selection:} reduce each cluster to a single robust representative by selecting the hypothesis closest to the cluster centroid.
    \item \textbf{Multi-view pose refinement:} translate the selected 3D pose to minimize multi-view reprojection error across all cameras.
\end{itemize}

For camera-constrained clustering we implement the following logic:
\begin{enumerate}
    \item Calculate $\mathcal{H}$ as $\mathcal{H}^c$ for all cameras $c$
    \item Create tuples $t=(id,c,\mathbf{P})$
    \item Start with a random tuple $t_1$ and assign it to cluster 1
    \item For all other tuples $t_i\in\{t_2, \dots,t_N\}$ and all clusters $\mathcal{C}^j$ with cluster centroid $\mathbf{k}^j$: if $||\mathbf{P}^i-\mathbf{k}^j||<\epsilon$ and not $c_i$ in $\mathcal{C}^j$ add $t_i$ to $\mathcal{C}^j$ else assign it to a new cluster   
\end{enumerate}

The camera constraint is crucial because, in crowded scenes, the intra-cluster variation can exceed the physical distance between nearby people (e.g., during physical contact). A purely distance-based clustering threshold would therefore either (i) be too small and split the same person into multiple clusters or (ii) be too large and merge two different people. By preventing two detections from the same camera from entering the same cluster, we can use a more permissive spatial threshold while substantially reducing the risk of merging distinct individuals.\par

After clustering, we must select a single pose per person. Since pose estimation can produce outliers, directly averaging all hypotheses may degrade quality. We instead compute the centroid of each cluster and choose the pose with minimal $\ell_2$ distance to this centroid (Figure~\ref{fig:multi_view_cluster}, right).\par

Once a representative pose is selected, we refine its 3D translation so that its joint projections agree with the 2D observations in all cameras. While absolute metric depth from a single view is less reliable (especially at long range), the 2D joint projections are accurate in most frames. We therefore minimize the multi-view reprojection error between the selected 3D pose and the per-camera \acrshort{sam}3 Body 3D outputs reprojected into the images, optimizing only the global translation using Nelder--Mead\cite{nelder1965simplex}.

\begin{figure}[ht!]
    \centering
    \includegraphics[width=\linewidth]{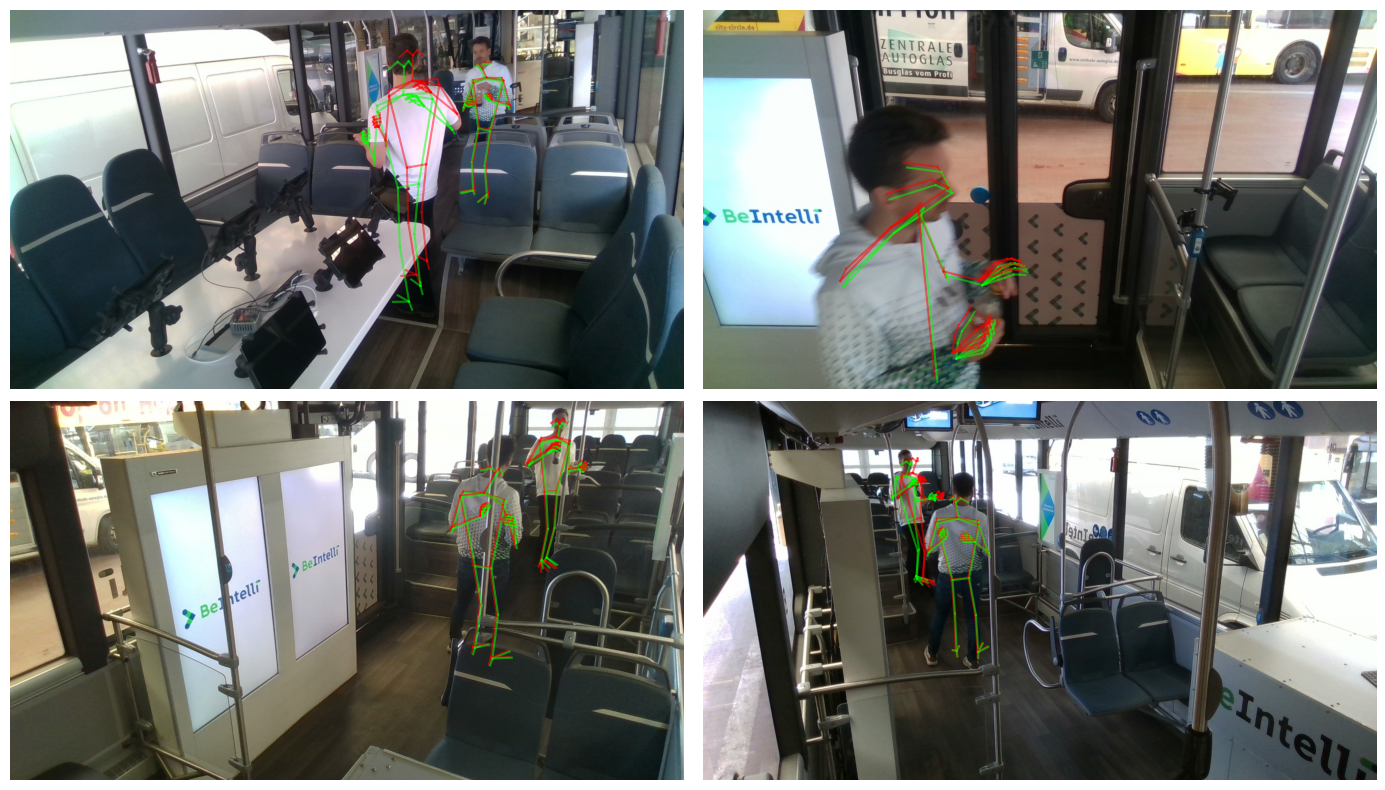}

    \caption{Refinement for projected poses from multiple views (\textcolor{red}{before} and \textcolor{green}{after})}
    \label{fig:pose_refinement}
\end{figure}

\begin{figure}[ht!]
    \centering
    \begin{subfigure}{0.48\linewidth}
        \includegraphics[width=\linewidth]{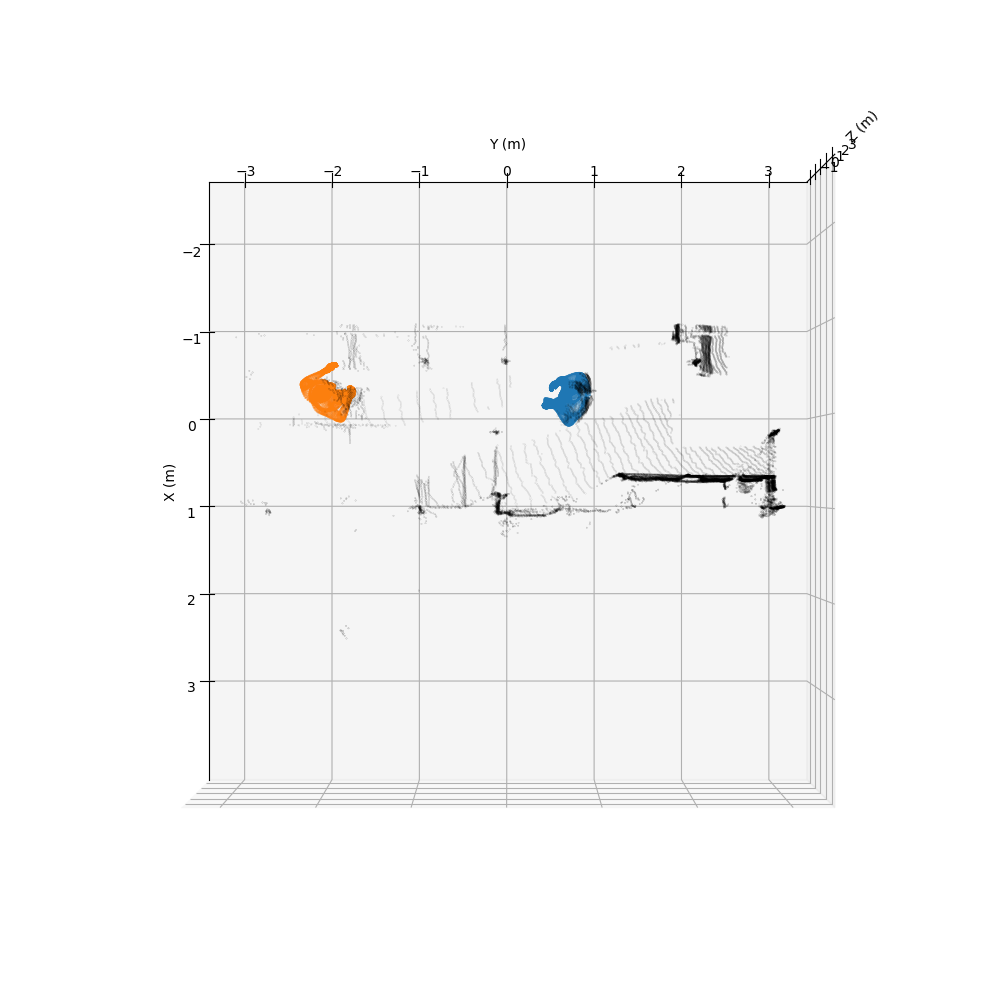}
        \label{fig:demo_lidar_mesh_ol1}
    \end{subfigure}
    \hfill
    \begin{subfigure}{0.48\linewidth}
        \includegraphics[width=\linewidth]{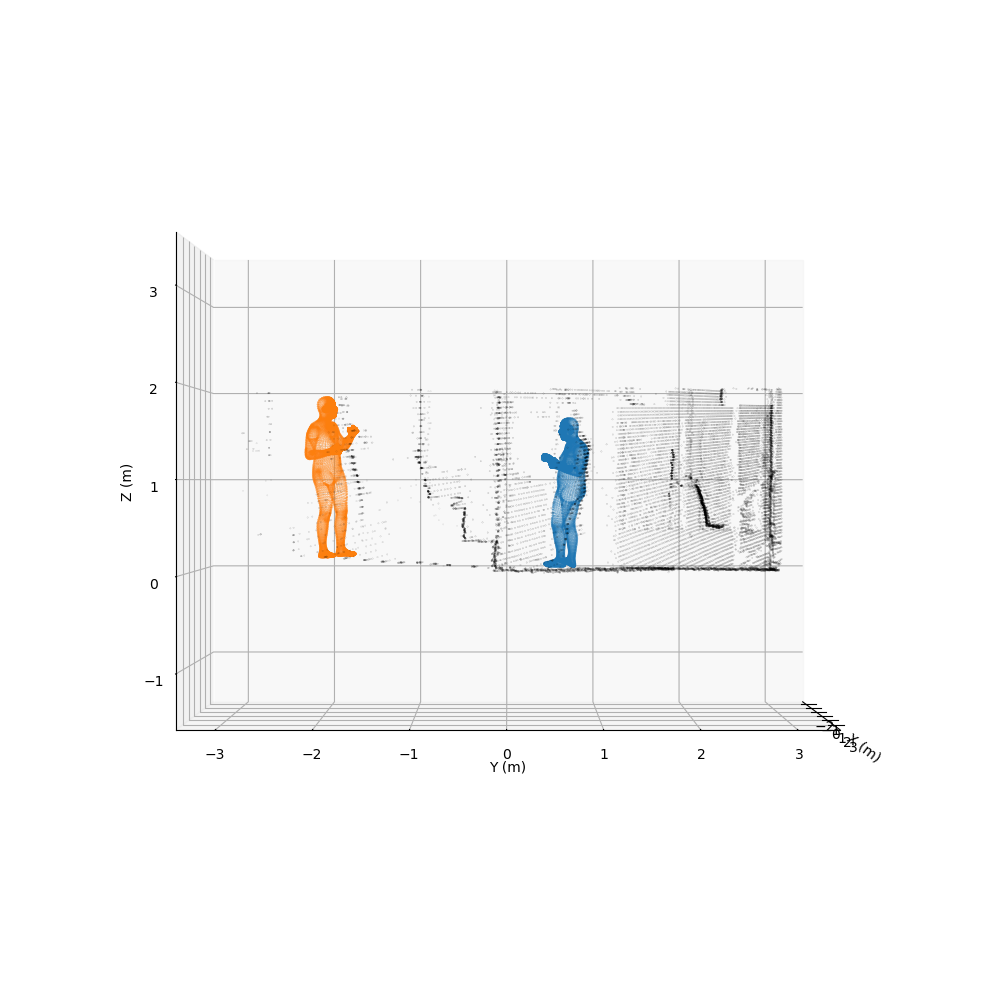}
        \label{fig:demo_lidar_mesh_ol2}
    \end{subfigure}

    \caption{LiDAR points with refined poses}
    \label{fig:demo_lidar_mesh_ol}
\end{figure}

We qualitatively validate metric accuracy of the refined 3D poses by transforming the according meshes into the LiDAR reference frame. As shown in Figure~\ref{fig:demo_lidar_mesh_ol}, the meshes align well with the LiDAR point cloud.\par

Finally, we convert each refined pose to a \acrshort{bev} bounding box and use these boxes as detections for a second ByteTrack pass, yielding a temporally consistent identity for each person. \par

As a final quality-control step, we manually curate the resulting tracks: we select the set of track identifiers corresponding to the instances we want to keep and remove remaining faulty detections that persist despite the multi-view aggregation. A visual review identified 48 cases across the dataset in which a pose wrongly assigned to cluster using the aforementioned aggregation procedure with $\epsilon=1$\,m.

\subsection{3D Bounding Box Extraction}

We next derive a 3D bounding box for every tracked person at each time step from the reconstructed body geometry. In line with standard autonomous-driving representations, each instance is encoded as an oriented 3D cuboid specified by its center $(x,y,z)$, side lengths $(l,w,h)$, and a yaw angle $\theta$.\par

Given a refined pose $\mathcal{P}^i_\tau$ at timestamp $\tau$, we compute the box by first expressing the pose points in a person-centric coordinate frame and then taking the axis-aligned bounds in that frame:

\begin{equation}
\begin{aligned} 
\mathcal{B}^i =
\operatorname{AABB}_3
\left(
\left\{
\mathbf{T}^{-1}
\begin{bmatrix}
\mathbf{p}\\
1
\end{bmatrix}
\,\middle|\,
\mathbf{p}\in\mathcal{P}^i_\tau
\right\}
\right).
\end{aligned}
\end{equation}

\begin{figure}[ht!]
    \centering
    \includegraphics[width=\linewidth]{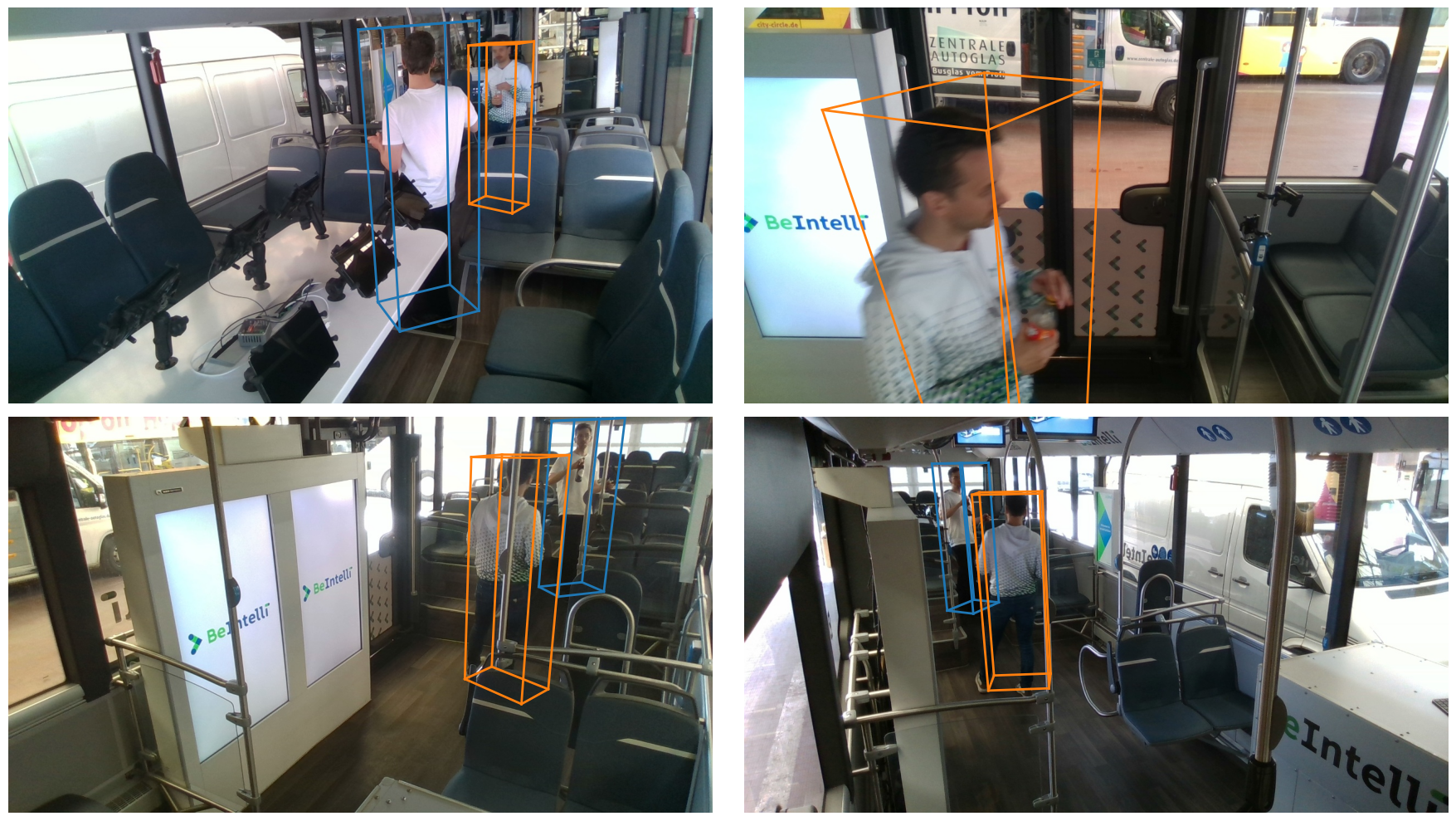}

    \caption{Example of 3D bounding box extraction from the generated 3D mesh.}
    \label{fig:bbox_extraction_demo}
\end{figure}

To obtain a canonical orientation, we define $\mathbf{T}\in\mathbb{R}^{4\times4}$ as the rigid transform from the world origin to a pelvis-centered frame whose origin is the midpoint between the left and right hip joints, and whose heading is given by the forward yaw direction at this point. Computing the bounds in this normalized frame yields a tight box whose orientation follows the subject's hip rotation. Figure~\ref{fig:bbox_extraction_demo} shows an example of the resulting 3D boxes.

\subsection{Action Recognition}\label{subsec:action_recognition}

In addition to the pseudo-labels for 3D pose and 3D bounding boxes, we provide an action label for each visible person at every timestamp. We distinguish between \,\emph{state} labels that describe the occupant's posture or locomotion (\textit{sit}, \textit{sit down}, \textit{stand up}, \textit{stand}, \textit{walk}, \textit{enter vehicle}, \textit{exit vehicle}, \textit{hold on}) and \,\emph{action} labels that capture semantically meaningful interactions or abnormal events (\textit{drinking non-/alcoholic beverages}, \textit{smoking}, \textit{defacement}, \textit{vandalism}, \textit{lost item}, \textit{mugging}, \textit{pushing}, \textit{littering}, \textit{punching}). The resulting taxonomy is inspired by prior in-cabin datasets \cite{tsiktsiris2020abnormal, tsiktsiris2025complete, ciampi2022bus, mishra2022cabin}.

\begin{figure}[ht!]
    \centering
    \begin{subfigure}{0.48\linewidth}
        \includegraphics[width=\linewidth]{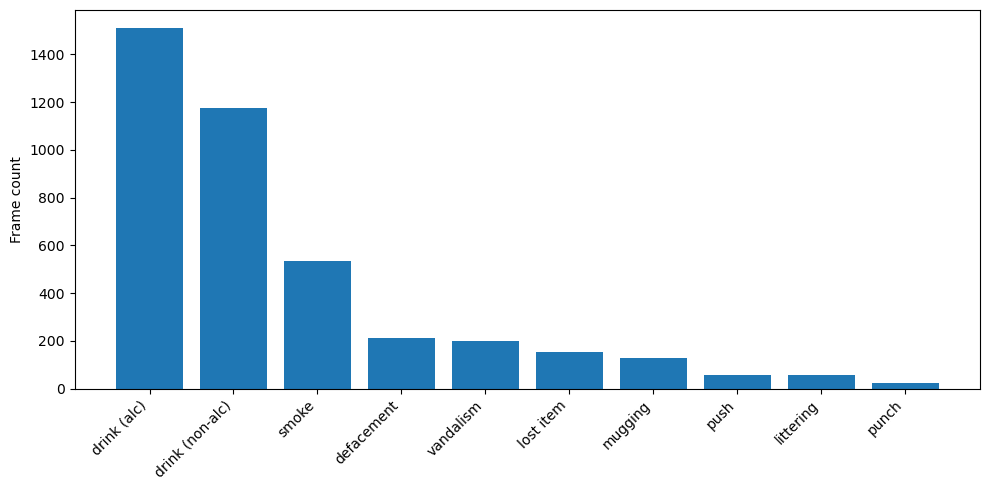}
        \caption{Frame count for actions}
    \end{subfigure}
    \begin{subfigure}{0.48\linewidth}
        \includegraphics[width=\linewidth]{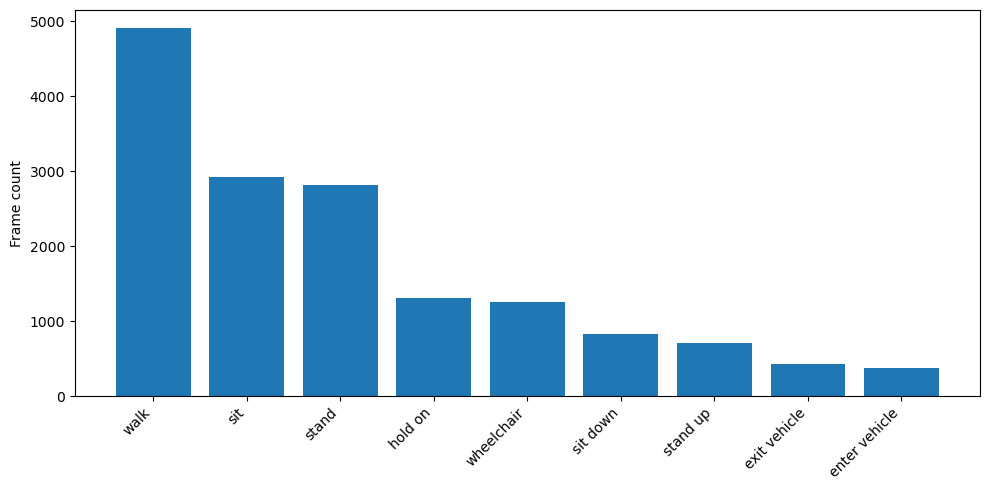}
        \caption{Frame count for states}
    \end{subfigure}
    \caption{Frame distribution for actions and states}
    \label{fig:action_state}
\end{figure}

Although these labels are not used in the experiments presented in this work, they support future \acrfull{har} studies in the proposed in-cabin setting \cite{sun2022human}. In contrast to large-scale \acrshort{har} benchmarks such as Kinetics-400 \cite{kay2017kinetics}, we intentionally keep the label set compact: our primary goal is accurate occupant localization, and the action annotations are provided as auxiliary supervision and for downstream analysis. Figure~\ref{fig:action_state} summarizes the label distribution, and qualitative examples are shown in Table~\ref{tab:action_recognition_examples}.

\begin{table}[t]
    \centering
    {\setlength{\tabcolsep}{3pt}% reduce inter-column padding to avoid overfull boxes
    \begin{tabular}{@{}ccc@{}}
        \parbox{0.3\linewidth}{\centering
            \includegraphics[width=0.9\linewidth]{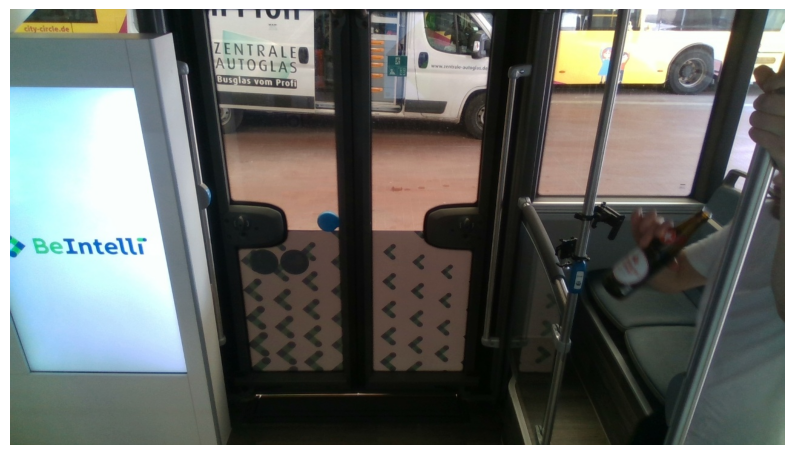}\\
            Drinking Alcohol
        } &
        \parbox{0.3\linewidth}{\centering
            \includegraphics[width=0.9\linewidth]{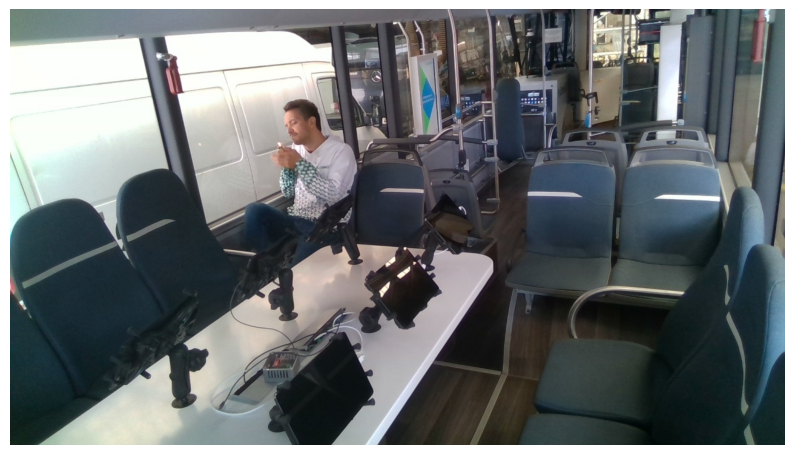}\\
            Smoking
        } &
        \parbox{0.3\linewidth}{\centering
            \includegraphics[width=0.9\linewidth]{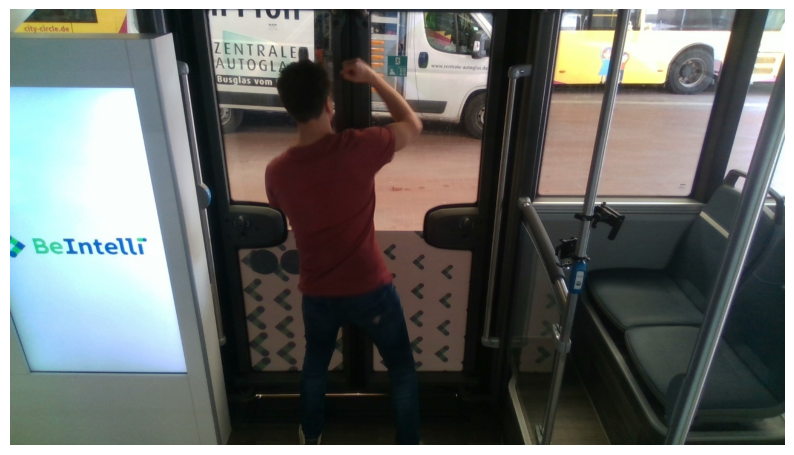}\\
            Vandalism
        } \\

        \parbox{0.3\linewidth}{\centering
            \includegraphics[width=0.9\linewidth]{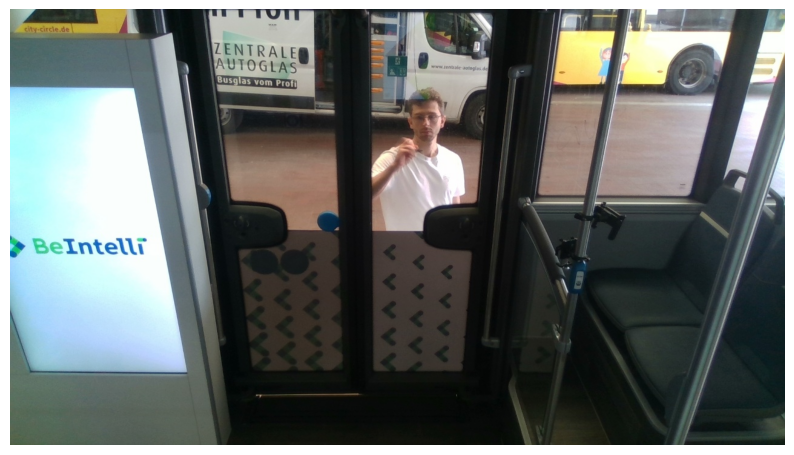}\\
            Defacement
        } &
        \parbox{0.3\linewidth}{\centering
            \includegraphics[width=0.9\linewidth]{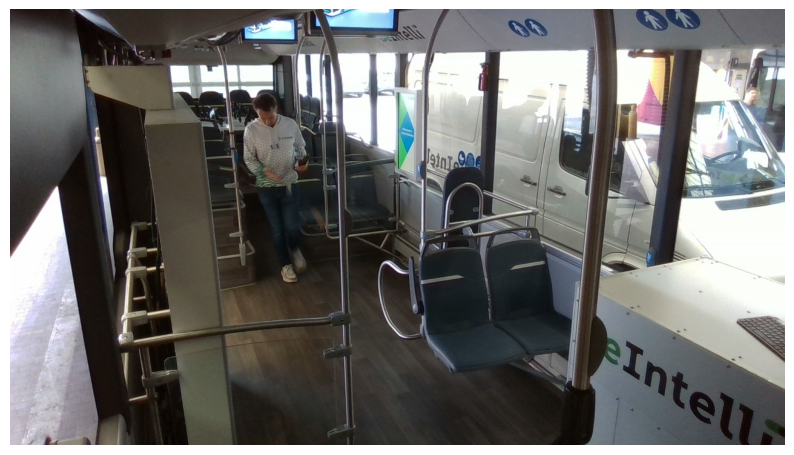}\\
            Littering
        } &
        \parbox{0.3\linewidth}{\centering
            \includegraphics[width=0.9\linewidth]{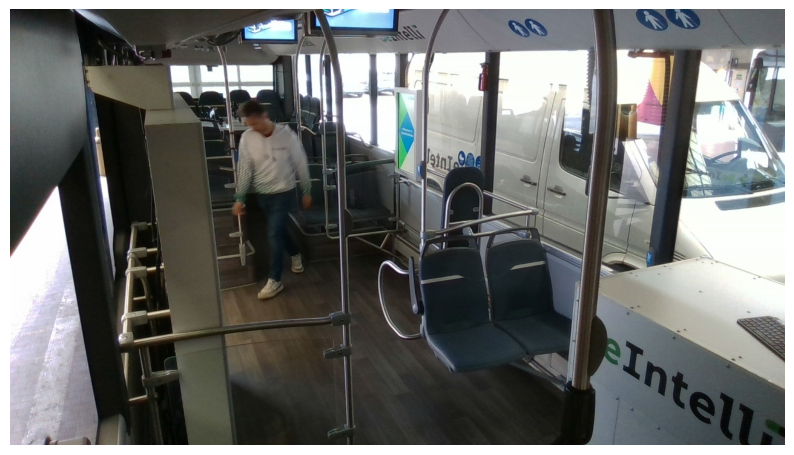}\\
            Lost Item
        } \\

        \parbox{0.3\linewidth}{\centering
            \includegraphics[width=0.9\linewidth]{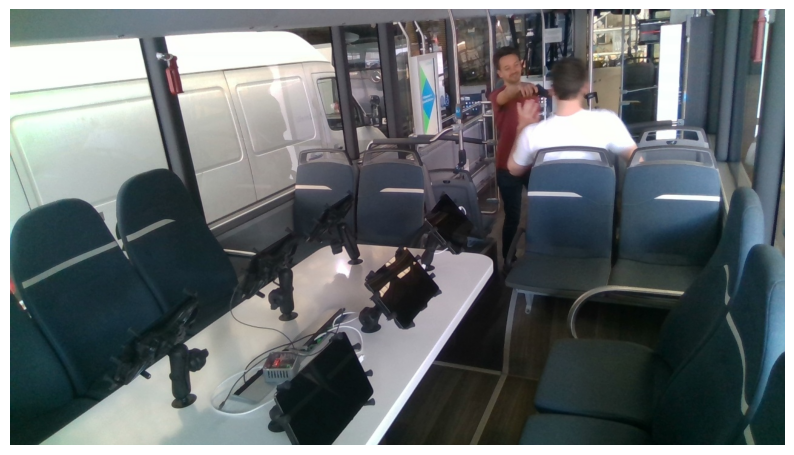}\\
            Armed Mugging
        } &
        \parbox{0.3\linewidth}{\centering
            \includegraphics[width=0.9\linewidth]{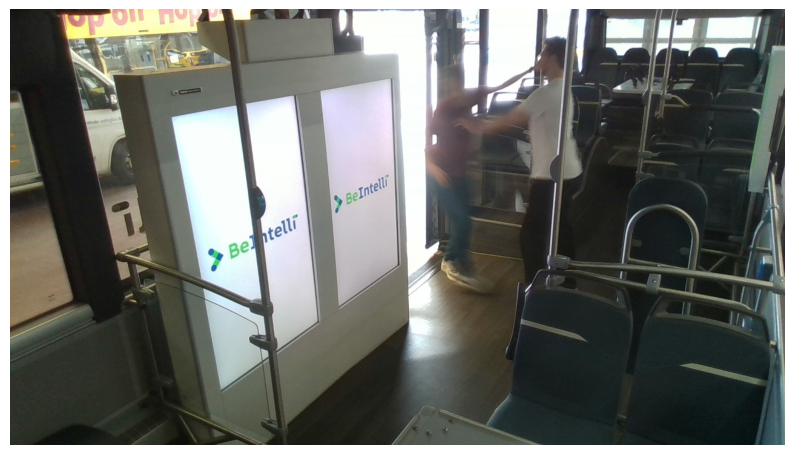}\\
            Violence
        } &
        \parbox{0.3\linewidth}{\centering
            \includegraphics[width=0.9\linewidth]{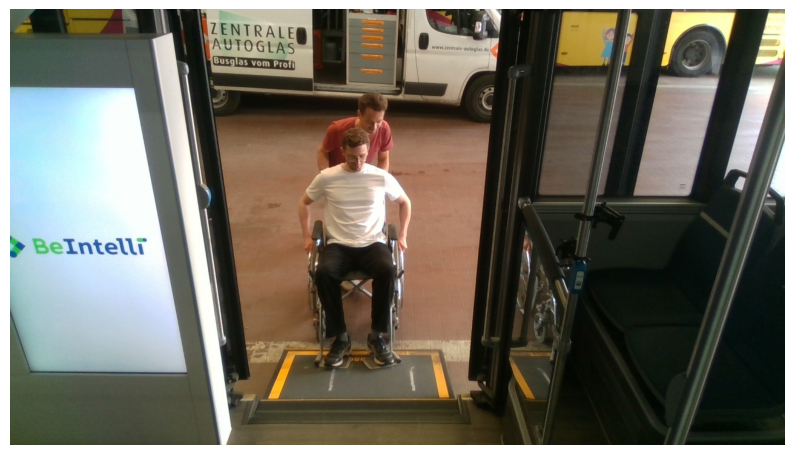}\\
            Wheelchair Assistance
        } \\
    \end{tabular}}

    \caption{Action recognition qualitative examples.}
    \label{tab:action_recognition_examples}
\end{table}

\section{Experiments}\label{experiments}

In this section, we evaluate state-of-the-art multi-view 3D object detection methods on the proposed in-cabin dataset. Although the dataset provides synchronized RGB and LiDAR, our primary focus is on camera-based, multi-view models that do not require LiDAR at inference time. \par

To ensure compatibility with existing codebases, we convert the data to the nuScenes format \cite{caesar2020nuscenes}. Because the vehicle remains stationary during data collection, we use a constant ego pose across frames. \par

We evaluate \acrfull{lss} \cite{philion2020lift} with two backbones: ResNet 50 \cite{he2016deep} and SwinT \cite{liu2021swin} as well as two variants of the VTransform module: \acrshort{lss}-Transform and AwareBEVDepth-Transform (\acrshort{abd}-Transform). In \acrshort{lss}-Transform, each camera image is resized and encoded into per-pixel features, and each pixel is ``lifted'' into a discrete set of depth bins, yielding a stack of frustum features that are then geometrically projected and aggregated into the \acrshort{bev} grid. \acrshort{abd}-Transform augments this lift step with explicit depth supervision: LiDAR points are projected into each image to form sparse depth labels, and an additional depth loss is minimized during training to encourage accurate per-pixel depth distributions while still performing camera-only inference at test time.

\begin{table}[ht!]
    \centering
    \begin{tabular}{lcccc}
        \hline\hline
        \textbf{Model} & \textbf{\acrshort{ap}}\textsubscript{0.2m} & \textbf{\acrshort{ap}}\textsubscript{0.5m} & \textbf{\acrshort{ap}}\textsubscript{1.0m} & \textbf{\acrshort{ap}}\textsubscript{1.5m} \\
        \hline
        ResNet & 0.7341 & 0.9630 & 0.9879 & 0.9880 \\
        ResNet-\acrshort{abd} & 0.6759 & 0.9397 & 0.9850 & 0.9853 \\
        SwinT & 0.3482 & 0.7563 & 0.8594 & 0.9218 \\
        SwinT-\acrshort{abd} & 0.3600 & 0.7334 & 0.8698 & 0.9391 \\
        \hline
        BEVFusion & 0.1870 & 0.7337 & 0.7549 & 0.7550 \\
        \hline\hline
    \end{tabular}
    \caption{Average precision (\acrshort{ap}~$\uparrow$) at different distance thresholds on the multi-view in-cabin dataset. }
    \label{tab:detection_results_ap}
\end{table}

\begin{figure}[ht!]
    \centering
    \includegraphics[width=0.6\linewidth]{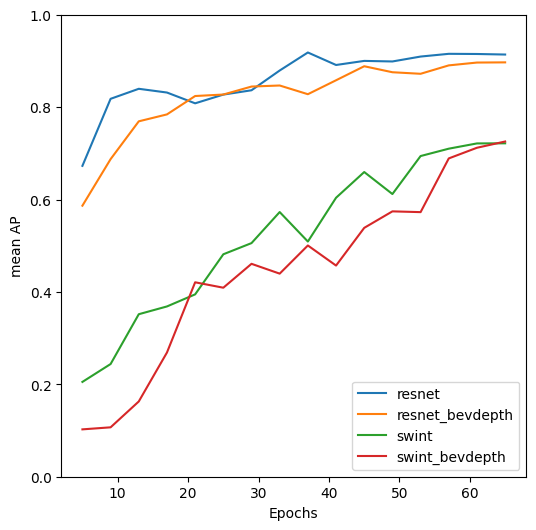}
    \caption{\acrshort{map} values on validation set}
    \label{fig:val_ap}
\end{figure}

Table~\ref{tab:detection_results_ap} summarizes the performance of the evaluated models on our multi-view in-cabin dataset. We remove samples without annotations and split the data into 80\%/20\% training/validation sets, resulting in 7{,}128 training samples and 2{,}008 validation samples.

We report standard nuScenes-style detection metrics: \acrfull{ap} at distance thresholds of 0.2~m, 0.5~m, 1.0~m, and 1.5~m, as well as \acrfull{ate}, \acrfull{ase}, and \acrfull{aoe}. We adapt the default nuScenes point-cloud range from \texttt{[-51.2, -51.2, -10, 51.2, 51.2, 10]}~m to \texttt{[-12.8, -12.8, -3, 12.8, 12.8, 3]}~m to match the in-cabin scale. All remaining hyperparameters follow the default configurations of the respective implementations. All \acrshort{lss} models were trained for 64 epochs, and the evolution of \acrshort{map} on the validation set is shown in Figure~\ref{fig:val_ap}. \par
 We additionally train \textbf{\acrshort{bev}Fusion} \cite{liu2023bevfusion}, a multi-modal fusion architecture that stacks LiDAR and image features in \acrshort{bev} space. Although the original \acrshort{bev}Fusion design is expected to outperform camera-only baselines, in our experiments the camera-only branch achieves higher performance than the multimodal variant. We attribute this discrepancy to the strong reliance of \acrshort{bev}Fusion on the LiDAR stream and to limitations of the SECOND \cite{yan2018second} backbone in the constrained vehicle-interior setting.

\begin{table}[t]
    \centering
    \begin{tabular}{l r}
        \hline
        \textbf{Item} & \textbf{Value} \\
        \hline
        Samples (\texttt{sample}) & 9{,}136 \\
        Sample data (\texttt{sample\_data}) & 45{,}680 \\
        Sample annotations (\texttt{sample\_annotation}) & 13{,}484 \\
        Total scenes & 10 \\
        Train / Val scenes & 8 / 2 \\
        Train / Val samples & 7{,}128 / 2{,}008 \\
        \hline
    \end{tabular}
n    \caption{nuScenes dataset statistics for in-cabin dataset}
    \label{tab:nuscenes_stats}
\end{table}

To characterize viewpoint-specific difficulty, we measure a missed-detection rate per-camera using Ultralytics YOLO26x \cite{Jocher_Ultralytics_YOLO_2023} in the complete evaluated dataset ($N=13484$ detections). Here, a miss is counted when a person is present in the scene but is not detected in a given camera view. 

\begin{table}[ht!]
    \centering
    \small
    \setlength{\tabcolsep}{6pt}
    \renewcommand{\arraystretch}{1.15}
    \begin{tabular}{lcc}
        \hline\hline
        \textbf{Camera} & \textbf{Miss rate} & \textbf{Misses/Total} \\
        \hline
        Back right & 24.28\% & 3274/13484 \\
        Center left & 60.69\% & 8183/13484 \\
        Front left & 18.78\% & 2532/13484 \\
        Front right & 31.20\% & 4207/13484 \\
        \hline\hline
    \end{tabular}
    \caption{Per-camera missed-detection rates on the complete evaluated dataset measured with Ultralytics YOLO26x \cite{Jocher_Ultralytics_YOLO_2023}.}
    \label{tab:missed_detections}
\end{table}

Table~\ref{tab:missed_detections} summarizes the resulting miss rates. Even at its best, a miss rate of $18.78\%$ implies that single-view detection is insufficient for vehicles at this scale, motivating a multi-view setup.
\section{Conclusion}\label{sec:conclusion}

We introduced a multi-view, multi-modal in-cabin monitoring dataset captured in a digitized city bus, together with a calibration and pseudo-labeling pipeline that generates 3D human pose estimates and oriented 3D bounding boxes for occupants. We also benchmarked representative multi-view 3D detection models on the resulting annotations to establish baselines for future work.

\paragraph{Calibration.}
We assess calibration quality at three levels. (i) The reconstructed cabin geometry is metrically consistent, as reflected by the marker edge-length statistics in Table~\ref{tab:edge-lengths}. (ii) Camera extrinsics remain stable under a PnP-based alignment, supporting consistent multi-view association. (iii) The LiDAR frame can be aligned to the COLMAP reconstruction via ICP, enabling cross-modal registration and reliable LiDAR-to-image reprojection.

\paragraph{Pseudo-label quality.}
Our multi-view aggregation strategy can separate multiple occupants while favoring temporally consistent tracks, but it is not failure-free: identities can still be merged or split, and the selected representative pose can be suboptimal under heavy occlusion. The subsequent multi-view refinement improves positional consistency across cameras.

\paragraph{Limitations and future work.}
Although the trained detectors achieve strong performance at larger matching thresholds (Table~\ref{tab:detection_results_ap}), performance at tight thresholds highlights remaining label noise and calibration sensitivity. Increasing dataset size and diversity (vehicles, sensor placements, illumination, and passenger demographics) is a key direction to improve generalization. Finally, while we provide per-frame action/state labels, training and benchmarking dedicated action-recognition models in this setting is left for future work.

\section*{Acknowledgment}
This work was conducted within the project ``Automation of Non-Driving Functions'', a PhD research collaboration between GT-ARC gemeinnützige GmbH, MAN Truck \& Bus SE, and TU Berlin. The project is funded and supervised by MAN and builds on resources developed in the government-funded BeIntelli project. In particular, we use the BeIntelli bus, which was procured and digitalized by GT-ARC and the DAI-Laboratory of TU Berlin with funding from the German Federal Ministry of Digital and Transport in the context of BeIntelli project.

\bibliographystyle{IEEEtran}
\bibliography{refs}

@inproceedings{leibe_pixelwise_2016,
  title={Pixelwise Instance Segmentation with a Dynamically Instantiated Network},
  author={Leibe, Bastian and others},
  booktitle={Proceedings of the IEEE Conference on Computer Vision and Pattern Recognition},
  year={2016},
  pages={123--132}
}

@article{mao20233d,
  title={3D object detection for autonomous driving: A comprehensive survey},
  author={Mao, Jiageng and Shi, Shaoshuai and Wang, Xiaogang and Li, Hongsheng},
  journal={International Journal of Computer Vision},
  volume={131},
  number={8},
  pages={1909--1963},
  year={2023},
  publisher={Springer}
}

@article{ming2021deep,
  title={Deep learning for monocular depth estimation: A review},
  author={Ming, Yue and Meng, Xuyang and Fan, Chunxiao and Yu, Hui},
  journal={Neurocomputing},
  volume={438},
  pages={14--33},
  year={2021},
  publisher={Elsevier}
}

@article{mishra2022cabin,
  title={In-cabin monitoring system for autonomous vehicles},
  author={Mishra, Ashutosh and Lee, Sangho and Kim, Dohyun and Kim, Shiho},
  journal={Sensors},
  volume={22},
  number={12},
  pages={4360},
  year={2022},
  publisher={MDPI}
}

@inproceedings{poon2022yolo,
  title={YOLO-based deep learning design for in-cabin monitoring system with fisheye-lens camera},
  author={Poon, Yen-Sok and Lin, Chih-Chun and Liu, Yu-Hsuan and Fan, Chih-Peng},
  booktitle={2022 IEEE International Conference on Consumer Electronics (ICCE)},
  pages={1--4},
  year={2022},
  organization={IEEE}
}

@inproceedings{mishra2020intelligent,
  title={An intelligent in-cabin monitoring system in fully autonomous vehicles},
  author={Mishra, Ashutosh and Kim, Jinhyuk and Kim, Dohyun and Cha, Jaekwang and Kim, Shiho},
  booktitle={2020 International SoC Design Conference (ISOCC)},
  pages={61--62},
  year={2020},
  organization={IEEE}
}

@article{sun2022human,
  title={Human action recognition from various data modalities: A review},
  author={Sun, Zehua and Ke, Qiuhong and Rahmani, Hossein and Bennamoun, Mohammed and Wang, Gang and Liu, Jun},
  journal={IEEE transactions on pattern analysis and machine intelligence},
  volume={45},
  number={3},
  pages={3200--3225},
  year={2022},
  publisher={IEEE}
}

@software{Jocher_Ultralytics_YOLO_2023,
  author = {Jocher, Glenn and Qiu, Jing and Chaurasia, Ayush},
  license = {AGPL-3.0},
  month = jan,
  title = {{Ultralytics YOLO}},
  url = {https://github.com/ultralytics/ultralytics},
  version = {8.0.0},
  year = {2023}
}

@article{tsiktsiris2025complete,
  title={A complete in-cabin monitoring framework for autonomous vehicles in public transportation},
  author={Tsiktsiris, Dimitris and Lalas, Antonios and Dasygenis, Minas and Votis, Konstantinos},
  journal={IET Intelligent Transport Systems},
  volume={19},
  number={1},
  pages={e12612},
  year={2025},
  publisher={Wiley Online Library}
}

@inproceedings{lin2024multi,
  title={Multi-modality action recognition based on dual feature shift in vehicle cabin monitoring},
  author={Lin, Dan and Lee, Philip Hann Yung and Li, Yiming and Wang, Ruoyu and Yap, Kim-Hui and Li, Bingbing and Ngim, You Shing},
  booktitle={ICASSP 2024-2024 IEEE International Conference on Acoustics, Speech and Signal Processing (ICASSP)},
  pages={6480--6484},
  year={2024},
  organization={IEEE}
}

@article{lin2024bushar,
  title={Abnormal activity detection and classification of bus passengers with in-vehicle image sensing},
  author={Lin, Huei-Yung and Tseng, Chun-Han},
  journal={IEEE Access},
  year={2024},
  doi={10.1109/ACCESS.2024.3365138}
}

@inproceedings{lin2014microsoft,
  title={Microsoft coco: Common objects in context},
  author={Lin, Tsung-Yi and Maire, Michael and Belongie, Serge and Hays, James and Perona, Pietro and Ramanan, Deva and Doll{\'a}r, Piotr and Zitnick, C Lawrence},
  booktitle={European conference on computer vision},
  pages={740--755},
  year={2014},
  organization={Springer}
}

@inproceedings{caesar2020nuscenes,
  title={nuScenes: A multimodal dataset for autonomous driving},
  author={Caesar, Holger and Bankiti, Varun and Lang, Alex H and Vora, Sourabh and Liong, Venice Erin and Xu, Qiang and Krishnan, Anush and Pan, Yu and Baldan, Giancarlo and Beijbom, Oscar},
  booktitle={Proceedings of the IEEE/CVF conference on computer vision and pattern recognition},
  pages={11621--11631},
  year={2020}
}

@article{geiger2013vision,
  title={Vision meets robotics: The kitti dataset},
  author={Geiger, Andreas and Lenz, Philip and Stiller, Christoph and Urtasun, Raquel},
  journal={The international journal of robotics research},
  volume={32},
  number={11},
  pages={1231--1237},
  year={2013},
  publisher={Sage Publications Sage UK: London, England}
}

@inproceedings{sun2020scalability,
  title={Scalability in perception for autonomous driving: Waymo open dataset},
  author={Sun, Pei and Kretzschmar, Henrik and Dotiwalla, Xerxes and Chouard, Aurelien and Patnaik, Vijaysai and Tsui, Paul and Guo, James and Zhou, Yin and Chai, Yuning and Caine, Benjamin and others},
  booktitle={Proceedings of the IEEE/CVF conference on computer vision and pattern recognition},
  pages={2446--2454},
  year={2020}
}

@article{koide2023general,
  title={General, single-shot, target-less, and automatic lidar-camera extrinsic calibration toolbox},
  author={Koide, Kenji and Oishi, Shuji and Yokozuka, Masashi and Banno, Atsuhiko},
  journal={arXiv preprint arXiv:2302.05094},
  year={2023}
}

@article{yan2022opencalib,
  title={Opencalib: A multi-sensor calibration toolbox for autonomous driving},
  author={Yan, Guohang and Liu, Zhuochun and Wang, Chengjie and Shi, Chunlei and Wei, Pengjin and Cai, Xinyu and Ma, Tao and Liu, Zhizheng and Zhong, Zebin and Liu, Yuqian and others},
  journal={Software Impacts},
  volume={14},
  pages={100393},
  year={2022},
  publisher={Elsevier}
}

@article{huang2020improvements,
  title={Improvements to target-based 3D LiDAR to camera calibration},
  author={Huang, Jiunn-Kai and Grizzle, Jessy W},
  journal={IEEE Access},
  volume={8},
  pages={134101--134110},
  year={2020},
  publisher={IEEE}
}

@inproceedings{schoenberger2016sfm,
    author={Sch\"{o}nberger, Johannes Lutz and Frahm, Jan-Michael},
    title={Structure-from-Motion Revisited},
    booktitle={Conference on Computer Vision and Pattern Recognition (CVPR)},
    year={2016},
}

@article{carion2025sam,
  title={Sam 3: Segment anything with concepts},
  author={Carion, Nicolas and Gustafson, Laura and Hu, Yuan-Ting and Debnath, Shoubhik and Hu, Ronghang and Suris, Didac and Ryali, Chaitanya and Alwala, Kalyan Vasudev and Khedr, Haitham and Huang, Andrew and others},
  journal={arXiv preprint arXiv:2511.16719},
  year={2025}
}

@article{chen1992object,
  title={Object modelling by registration of multiple range images},
  author={Chen, Yang and Medioni, G{\'e}rard},
  journal={Image and vision computing},
  volume={10},
  number={3},
  pages={145--155},
  year={1992},
  publisher={Elsevier}
}

@inproceedings{maji2022yolo,
  title={Yolo-pose: Enhancing yolo for multi person pose estimation using object keypoint similarity loss},
  author={Maji, Debapriya and Nagori, Soyeb and Mathew, Manu and Poddar, Deepak},
  booktitle={Proceedings of the IEEE/CVF conference on computer vision and pattern recognition},
  pages={2637--2646},
  year={2022}
}

@inproceedings{zhang2022bytetrack,
  title={Bytetrack: Multi-object tracking by associating every detection box},
  author={Zhang, Yifu and Sun, Peize and Jiang, Yi and Yu, Dongdong and Weng, Fucheng and Yuan, Zehuan and Luo, Ping and Liu, Wenyu and Wang, Xinggang},
  booktitle={European conference on computer vision},
  pages={1--21},
  year={2022},
  organization={Springer}
}

@article{li2024bevformer,
  title={Bevformer: learning bird’s-eye-view representation from lidar-camera via spatiotemporal transformers},
  author={Li, Zhiqi and Wang, Wenhai and Li, Hongyang and Xie, Enze and Sima, Chonghao and Lu, Tong and Yu, Qiao and Dai, Jifeng},
  journal={IEEE Transactions on Pattern Analysis and Machine Intelligence},
  volume={47},
  number={3},
  pages={2020--2036},
  year={2024},
  publisher={IEEE}
}

@inproceedings{philion2020lift,
  title={Lift, splat, shoot: Encoding images from arbitrary camera rigs by implicitly unprojecting to 3d},
  author={Philion, Jonah and Fidler, Sanja},
  booktitle={European conference on computer vision},
  pages={194--210},
  year={2020},
  organization={Springer}
}

@inproceedings{liu2023bevfusion,
  title={Bevfusion: Multi-task multi-sensor fusion with unified bird's-eye view representation},
  author={Liu, Zhijian and Tang, Haotian and Amini, Alexander and Yang, Xinyu and Mao, Huizi and Rus, Daniela L and Han, Song},
  booktitle={2023 IEEE international conference on robotics and automation (ICRA)},
  pages={2774--2781},
  year={2023},
  organization={IEEE}
}

@inproceedings{joo2015panoptic,
  title={Panoptic studio: A massively multiview system for social motion capture},
  author={Joo, Hanbyul and Liu, Hao and Tan, Lei and Gui, Lin and Nabbe, Bart and Matthews, Iain and Kanade, Takeo and Nobuhara, Shohei and Sheikh, Yaser},
  booktitle={Proceedings of the IEEE international conference on computer vision},
  pages={3334--3342},
  year={2015}
}

@article{ionescu2013human3,
  title={Human3. 6m: Large scale datasets and predictive methods for 3d human sensing in natural environments},
  author={Ionescu, Catalin and Papava, Dragos and Olaru, Vlad and Sminchisescu, Cristian},
  journal={IEEE transactions on pattern analysis and machine intelligence},
  volume={36},
  number={7},
  pages={1325--1339},
  year={2013},
  publisher={IEEE}
}

@InProceedings{Srivastav_2024_CVPR,
    author    = {Srivastav, Vinkle and Chen, Keqi and Padoy, Nicolas},
    title     = {SelfPose3d: Self-Supervised Multi-Person Multi-View 3d Pose Estimation},
    booktitle = {Proceedings of the IEEE/CVF Conference on Computer Vision and Pattern Recognition (CVPR)},
    month     = {June},
    year      = {2024},
    pages     = {2502-2512}
}

@article{zheng20213d,
title={3D Human Pose Estimation with Spatial and Temporal Transformers},
author={Zheng, Ce and Zhu, Sijie and Mendieta, Matias and Yang, Taojiannan and Chen, Chen and Ding, Zhengming},
journal={Proceedings of the IEEE International Conference on Computer Vision (ICCV)},
year={2021}
}

@inproceedings{chang2019argoverse,
  title={Argoverse: 3d tracking and forecasting with rich maps},
  author={Chang, Ming-Fang and Lambert, John and Sangkloy, Patsorn and Singh, Jagjeet and Bak, Slawomir and Hartnett, Andrew and Wang, De and Carr, Peter and Lucey, Simon and Ramanan, Deva and others},
  booktitle={Proceedings of the IEEE/CVF conference on computer vision and pattern recognition},
  pages={8748--8757},
  year={2019}
}

@article{wilson2023argoverse,
  title={Argoverse 2: Next generation datasets for self-driving perception and forecasting},
  author={Wilson, Benjamin and Qi, William and Agarwal, Tanmay and Lambert, John and Singh, Jagjeet and Khandelwal, Siddhesh and Pan, Bowen and Kumar, Ratnesh and Hartnett, Andrew and Pontes, Jhony Kaesemodel and others},
  journal={arXiv preprint arXiv:2301.00493},
  year={2023}
}

@inproceedings{andriluka20142d,
  title={2d human pose estimation: New benchmark and state of the art analysis},
  author={Andriluka, Mykhaylo and Pishchulin, Leonid and Gehler, Peter and Schiele, Bernt},
  booktitle={Proceedings of the IEEE Conference on computer Vision and Pattern Recognition},
  pages={3686--3693},
  year={2014}
}

@article{valverde2025survey,
  title={A Survey of Deep Learning-Based 3D Object Detection Methods for Autonomous Driving Across Different Sensor Modalities},
  author={Valverde, Miguel and Moutinho, Alexandra and Zacchi, Jo{\~a}o-Vitor},
  journal={Sensors (Basel, Switzerland)},
  volume={25},
  number={17},
  pages={5264},
  year={2025}
}

@article{yang2026sam,
  title={Sam 3d body: Robust full-body human mesh recovery},
  author={Yang, Xitong and Kukreja, Devansh and Pinkus, Don and Sagar, Anushka and Fan, Taosha and Park, Jinhyung and Shin, Soyong and Cao, Jinkun and Liu, Jiawei and Ugrinovic, Nicolas and others},
  journal={arXiv preprint arXiv:2602.15989},
  year={2026}
}

@article{hinton2015distilling,
  title={Distilling the knowledge in a neural network},
  author={Hinton, Geoffrey and Vinyals, Oriol and Dean, Jeff},
  journal={arXiv preprint arXiv:1503.02531},
  year={2015}
}

@article{yalniz2019billion,
  title={Billion-scale semi-supervised learning for image classification},
  author={Yalniz, I Zeki and J{\'e}gou, Herv{\'e} and Chen, Kan and Paluri, Manohar and Mahajan, Dhruv},
  journal={arXiv preprint arXiv:1905.00546},
  year={2019}
}

@article{ciampi2022bus,
  title={Bus violence: An open benchmark for video violence detection on public transport},
  author={Ciampi, Luca and Foszner, Pawe{\l} and Messina, Nicola and Staniszewski, Micha{\l} and Gennaro, Claudio and Falchi, Fabrizio and Serao, Gianluca and Cogiel, Micha{\l} and Golba, Dominik and Szczesna, Agnieszka and others},
  journal={Sensors},
  volume={22},
  number={21},
  pages={8345},
  year={2022},
  publisher={MDPI}
}

@inproceedings{xie2020self,
  title={Self-training with noisy student improves imagenet classification},
  author={Xie, Qizhe and Luong, Minh-Thang and Hovy, Eduard and Le, Quoc V},
  booktitle={Proceedings of the IEEE/CVF conference on computer vision and pattern recognition},
  pages={10687--10698},
  year={2020}
}

@article{french2017self,
  title={Self-ensembling for visual domain adaptation},
  author={French, Geoffrey and Mackiewicz, Michal and Fisher, Mark},
  journal={arXiv preprint arXiv:1706.05208},
  year={2017}
}

@inproceedings{lee2013pseudo,
  title={Pseudo-label: The simple and efficient semi-supervised learning method for deep neural networks},
  author={Lee, Dong-Hyun and others},
  booktitle={Workshop on challenges in representation learning, ICML},
  volume={3},
  pages={896},
  year={2013},
  organization={Atlanta}
}

@inproceedings{meyer2023cherrypicker,
  title={CherryPicker: Semantic skeletonization and topological reconstruction of cherry trees},
  author={Meyer, Lukas and Gilson, Andreas and Scholz, Oliver and Stamminger, Marc},
  booktitle={Proceedings of the IEEE/CVF Conference on Computer Vision and Pattern Recognition},
  pages={6244--6253},
  year={2023}
}

@Article{tsiktsiris2020abnormal,
AUTHOR = {Tsiktsiris, Dimitris and Dimitriou, Nikolaos and Lalas, Antonios and Dasygenis, Minas and Votis, Konstantinos and Tzovaras, Dimitrios},
TITLE = {Real-Time Abnormal Event Detection for Enhanced Security in Autonomous Shuttles Mobility Infrastructures},
JOURNAL = {Sensors},
VOLUME = {20},
YEAR = {2020},
NUMBER = {17},
ARTICLE-NUMBER = {4943},
URL = {https://www.mdpi.com/1424-8220/20/17/4943},
PubMedID = {32882846},
ISSN = {1424-8220},
DOI = {10.3390/s20174943}
}

@ARTICLE{tsiktsiris2024overhead,
  author={Tsiktsiris, Dimitris and Lalas, Antonios and Dasygenis, Minas and Votis, Konstantinos},
  journal={IEEE Access}, 
  title={Improving Passenger Detection With Overhead Fisheye Imaging}, 
  year={2024},
  volume={12},
  number={},
  pages={66237-66247},
  doi={10.1109/ACCESS.2024.3395786}}

@article{kay2017kinetics,
  title={The kinetics human action video dataset},
  author={Kay, Will and Carreira, Joao and Simonyan, Karen and Zhang, Brian and Hillier, Chloe and Vijayanarasimhan, Sudheendra and Viola, Fabio and Green, Tim and Back, Trevor and Natsev, Paul and others},
  journal={arXiv preprint arXiv:1705.06950},
  year={2017}
}

@inproceedings{he2016deep,
  title={Deep residual learning for image recognition},
  author={He, Kaiming and Zhang, Xiangyu and Ren, Shaoqing and Sun, Jian},
  booktitle={Proceedings of the IEEE conference on computer vision and pattern recognition},
  pages={770--778},
  year={2016}
}

@inproceedings{liu2021swin,
  title={Swin transformer: Hierarchical vision transformer using shifted windows},
  author={Liu, Ze and Lin, Yutong and Cao, Yue and Hu, Han and Wei, Yixuan and Zhang, Zheng and Lin, Stephen and Guo, Baining},
  booktitle={Proceedings of the IEEE/CVF international conference on computer vision},
  pages={10012--10022},
  year={2021}
}

@article{yan2018second,
  title={Second: Sparsely embedded convolutional detection},
  author={Yan, Yan and Mao, Yuxing and Li, Bo},
  journal={Sensors},
  volume={18},
  number={10},
  pages={3337},
  year={2018},
  publisher={MDPI}
}

@article{nelder1965simplex,
  title={A simplex method for function minimization},
  author={Nelder, John A and Mead, Roger},
  journal={The computer journal},
  volume={7},
  number={4},
  pages={308--313},
  year={1965},
  publisher={The British Computer Society}
}

\end{document}